\documentclass[10pt,twocolumn,letterpaper]{article}

\usepackage{iccv}
\usepackage{times}
\usepackage{epsfig}

\usepackage{graphicx}
\usepackage{amsmath}
\usepackage{amssymb}
\usepackage{booktabs}
\usepackage{subcaption}

\usepackage{multirow, array, tabularx, amsfonts, bm, xcolor}
\usepackage{adjustbox}
\usepackage{arydshln}
\usepackage{pifont}
\usepackage{capt-of}
\usepackage{float,wrapfig}
\usepackage{comment}
\usepackage{array}
\usepackage{algorithm}
\usepackage{algpseudocode}
\usepackage{url}
\usepackage{mathtools, empheq}
\usepackage{enumitem}

\newcommand{\cmark}{\ding{51}} 
\newcommand{\xmark}{\ding{55}} 

\newcolumntype{P}[1]{>{\centering\arraybackslash}p{#1}}

\algnewcommand\algorithmicforeach{\textbf{for each}}
\algdef{S}[FOR]{ForEach}[1]{\algorithmicforeach\ #1\ \algorithmicdo}

\newcommand{\green}[1]{{\color{green} #1}}
\newcommand{\red}[1]{{\color{red} #1}}

\usepackage[pagebackref=true,breaklinks=true,letterpaper=true,colorlinks,bookmarks=false]{hyperref}

\iccvfinalcopy 

\def\iccvPaperID{9728} 

\ificcvfinal\fi

\begin{document}

\title{Rethinking the Role of Pre-Trained Networks \\ in Source-Free Domain Adaptation}

\author{Wenyu Zhang$^1$, Li Shen$^1$, Chuan-Sheng Foo$^{1,2}$\\
$^1$Institute for Infocomm Research (I$^\text{2}$R), Agency for Science, Technology and Research (A*STAR)\\
$^2$Centre for Frontier AI Research (CFAR), Agency for Science, Technology and Research (A*STAR)\\
}

\maketitle
\ificcvfinal\thispagestyle{empty}\fi

\begin{abstract}
Source-free domain adaptation (SFDA) aims to adapt a source model trained on a fully-labeled source domain to an unlabeled target domain. Large-data pre-trained networks are used to initialize source models during source training, and subsequently discarded. However, source training can cause the model to overfit to source data distribution and lose applicable target domain knowledge. 
We propose to integrate the pre-trained network into the target adaptation process as it has diversified features important for generalization and provides an alternate view of features and classification decisions different from the source model.
We propose to distil useful target domain information through a co-learning strategy to improve target pseudolabel quality for finetuning the source model. Evaluation on 4 benchmark datasets show that our proposed strategy improves adaptation performance and can be successfully integrated with existing SFDA methods. Leveraging modern pre-trained networks that have stronger representation learning ability in the co-learning strategy further boosts performance.

\end{abstract}

\section{Introduction}
\label{sec: introduction}

Deep neural networks have demonstrated remarkable ability in diverse applications, but their performance typically relies on the assumption that training \emph{(source domain)} and test \emph{(target domain)} data distributions are the same. This assumption can be violated in practice when the source data does not fully represent the entire data distribution due to difficulties of real-world data collection. Target samples distributed differently from source samples (due to factors such as background, illumination and style~\cite{gulrajani2020domainbed,wilds}) are subject to \emph{domain shift} (also known as \emph{covariate shift}), and can severely degrade model performance.

\emph{Domain adaptation} (DA) aims to tackle the domain shift problem by transferring knowledge from a fully-labeled source domain to an unlabeled target domain. The classic setting of \emph{unsupervised domain adaptation} assumes source and target data are jointly available for training~\cite{wilson2020dasurvey}. Motivated by the theoretical bound on target risk derived in \cite{BenDavid2010domainadaptation}, a key strategy is to minimize the discrepancy between source and target features to learn domain-invariant features~\cite{Ganin2016DANN,Li2018CDANN,li2018mmd,Sun2016DeepCORAL, zhang2019mdd,hu2020gsda,kang2019can,gu2020rsda,Xu2019sfan}. However, access to source data can be impractical due to data privacy concerns. Recently, the \emph{source-free domain adaptation} (SFDA) setting is proposed~\cite{liang2020shot} to split adaptation into two stages: (a) training network with fully labeled source data, and (b) adapting source model with unlabeled target data. An example use case in a corporate context is when the vendor has collected data to train a source model, and clients seek to address the same task for their own environments, but data sharing for joint training cannot be achieved due to proprietary or privacy reasons.
The vendor makes available the source model, which the clients can adapt with available resources. In this work, we assume the role of the clients.

\begin{figure}
  \centering
  \begin{subfigure}{0.48\linewidth}
    \includegraphics[width=\linewidth]{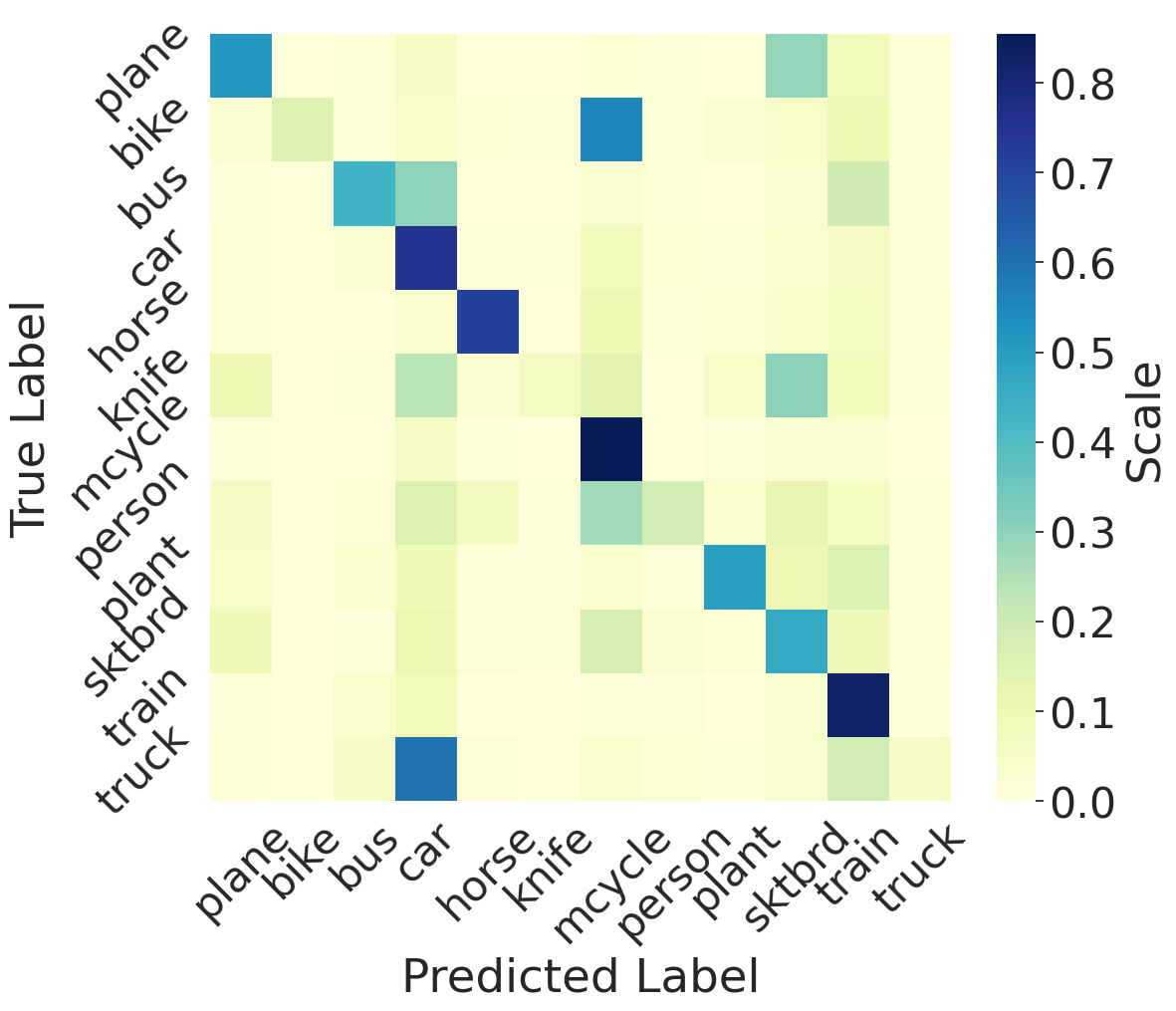}
    \caption{Confusion matrix}
    \label{fig: visda_confusion_matrix}
  \end{subfigure}
  \hfill
  \begin{subfigure}{0.48\linewidth}
    \includegraphics[width=\linewidth]{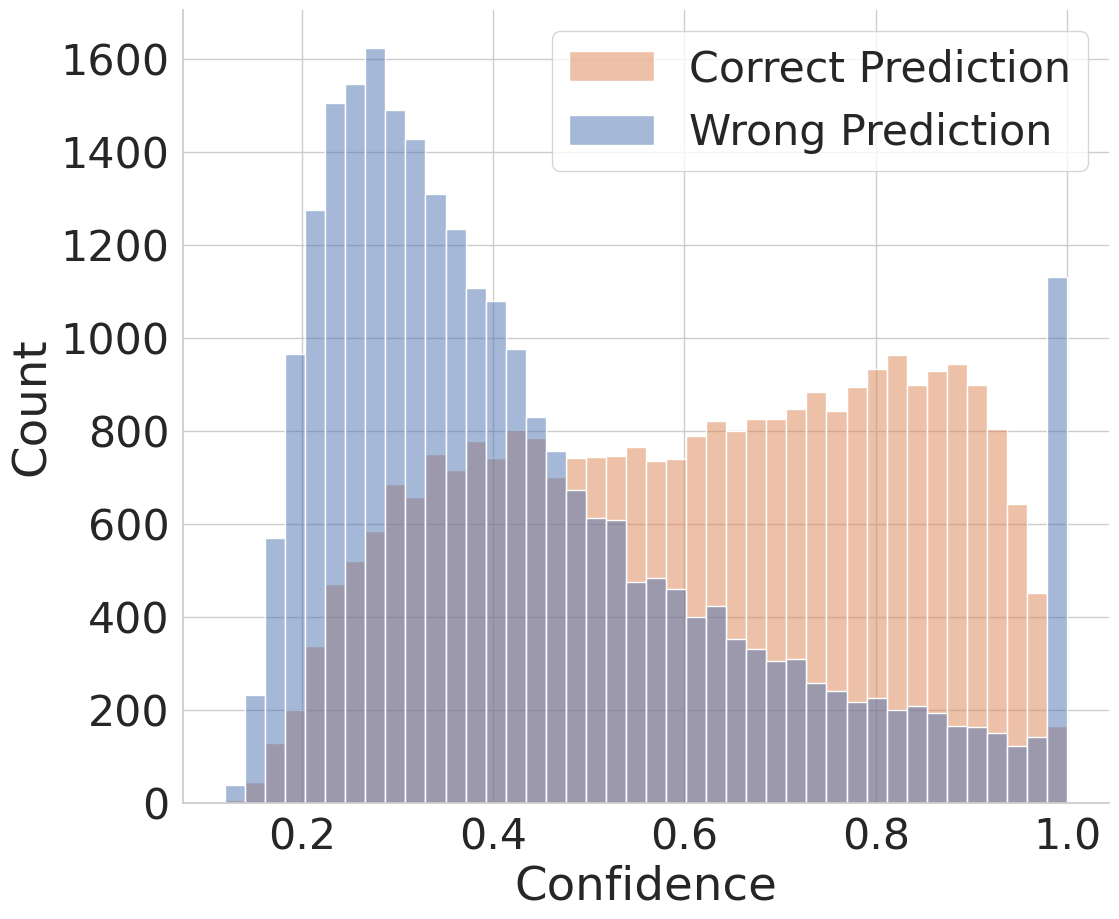}
    \caption{Distribution of confidence}
    \label{fig: visda_confidence}
  \end{subfigure}
  
  \vspace{-2mm}
  \caption{VisDA-C source-trained ResNet-101 produces unreliable pseudolabels on target samples, and is over-confident on a significant number of incorrect predictions.}
  \label{fig: visda_prediction}
  \vspace{-4mm}
\end{figure}

We focus on the classification task where source and target domain share the same label space.
The source model is typically obtained by training a selected network on source data with supervised loss.  
For adaptation, existing SFDA methods generate or estimate source-like representations to align source and target distributions~\cite{qiu2021prototypegen,ding2022sfdade}, make use of local clustering structures in the target data~\cite{yang2021gsfda,Yang2021ExploitingTI,Yang2022AttractingAD}, and learn semantics through self-supervised tasks~\cite{xia2021a2net,liang2021shotplus}.
\cite{liang2020shot,Kundu2020TowardsIM,li20203cgan} use the source model to generate target pseudolabels for finetuning, and \cite{kim2021sfda,chen2021noisy,liang2021shotplus} further select samples with the low-entropy or low-loss criterion. However, model calibration is known to degrade under distribution shift~\cite{ovadia2019uncertainty}. We observe in Figure~\ref{fig: visda_prediction} that target pseudolabels produced by source model can be biased, and outputs such as prediction confidence (and consequently entropy and loss) may not reflect accuracy and cannot reliably be used alone to improve pseudolabel quality.

\begin{figure}[tb]
    \centering
    \includegraphics[width=\linewidth]{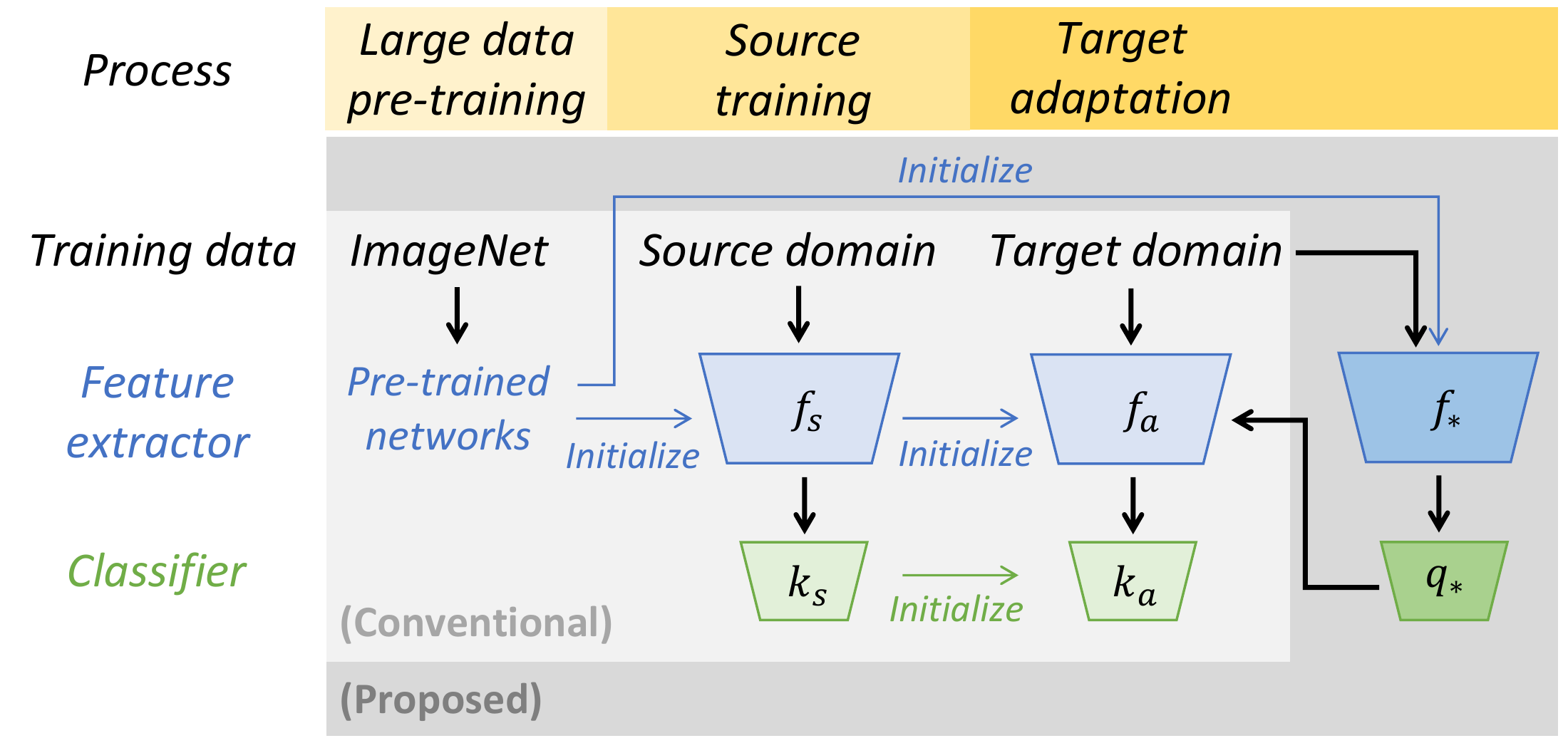}
    
    \vspace{-2mm}
    \caption{Overview of conventional and proposed frameworks: We propose incorporating pre-trained networks during target adaptation.}
    \label{fig: overview_setting}
    \vspace{-2mm}
\end{figure}

We reconsider the role of pre-trained networks in SFDA. Due to their common usage, the discussion in this paper is within the scope of ImageNet networks, but can be applied to other large data pre-trained networks. As in Figure~\ref{fig: overview_setting}, ImageNet weights are conventionally used to initialize source models and subsequently discarded. However, ImageNet networks have diverse features important for generalization~\cite{chen2021contrastive}, and finetuning on source data can overfit to source distribution and potentially lose pre-existing target information. Furthermore, modern ImageNet networks may have better generalizability following new architectures and training schemes. 
We seek to answer the questions:
\begin{itemize}[noitemsep]
    \item \emph{Can finetuning pre-trained networks on source data lose applicable target domain knowledge?}
    \item \emph{Given a source model, whether and how pre-trained networks can help its adaptation?}
\end{itemize}

\begin{figure}[tb]
    \centering
    \includegraphics[width=\linewidth]{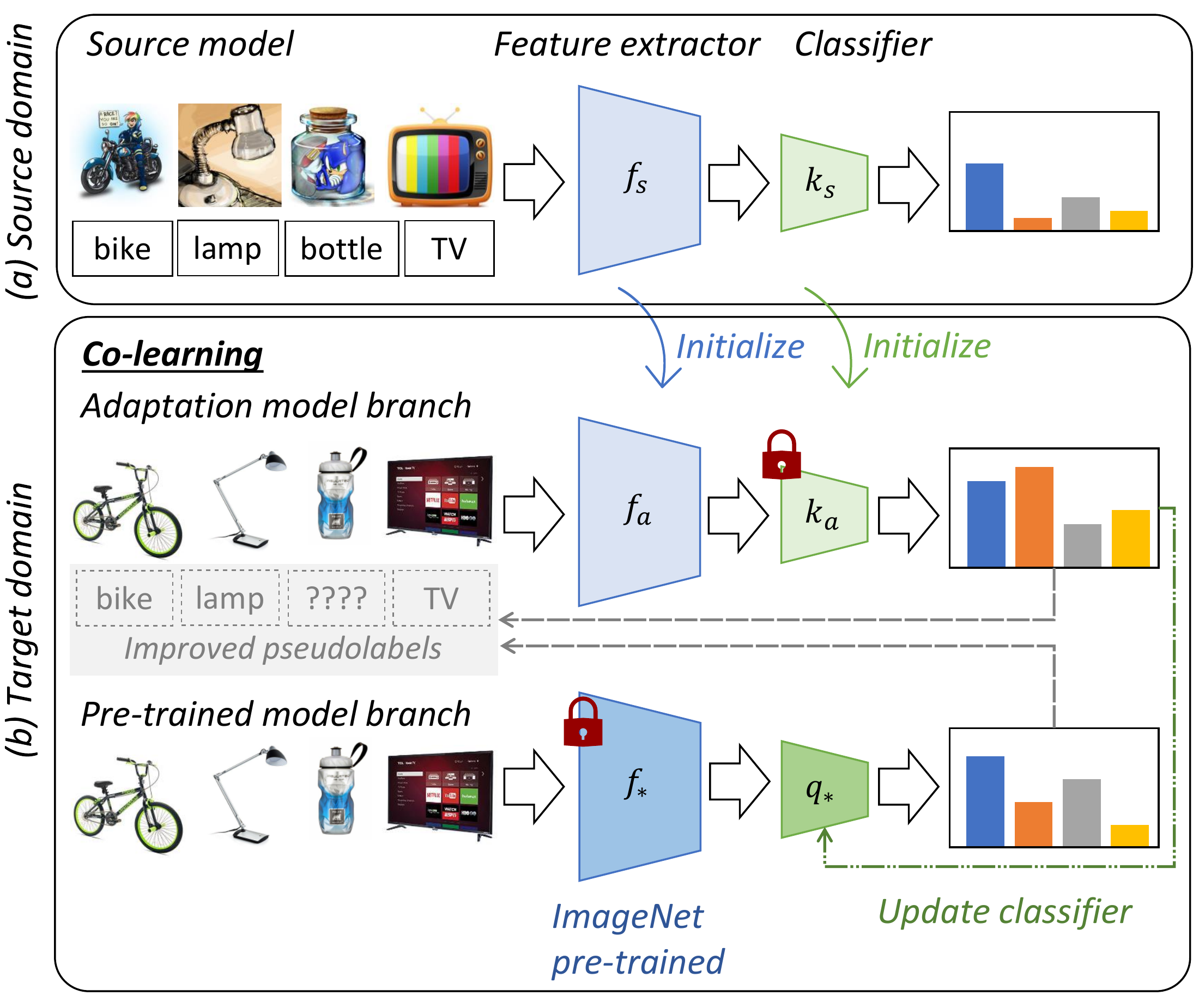}
    
    \vspace{-2mm}
    \caption{Overview of proposed strategy: (a) Source model trained on source domain is provided. (b) We adapt the source model through a co-learning strategy where the adaptation model and an ImageNet pre-trained model collectively produce more reliable pseudolabels for finetuning. Note since the ImageNet feature extractor is frozen, the strategy can be implemented with a feature bank.}
    \label{fig: overview}
    \vspace{-4mm}
\end{figure}

We propose to integrate ImageNet networks into the target adaptation process, as depicted in Figure~\ref{fig: overview_setting}, to distil any effective target domain knowledge they may hold. They can help to correct source model prediction bias and produce more accurate target pseudolabels to finetune the source model. We design a simple two-branch co-learning strategy where the adaptation and pre-trained model branch iteratively updates to collaboratively generate more accurate pseudolabels, which can be readily applied to existing SFDA methods as well. We provide an overview of our strategy in Figure~\ref{fig: overview}, and summarize our contributions:
\begin{itemize}[noitemsep]
    \item We observe that finetuning pre-trained networks on source data can lose generalizability for target domain;
    \item Based on the above observation, we propose integrating pre-trained networks into SFDA target adaptation;
    \item We propose a simple co-learning strategy to distil useful target domain information from a pre-trained feature extractor to improve target pseudolabel quality;
    \item We demonstrate performance improvements by the proposed strategy (including just reusing the ImageNet network in source model initialization) on 4 benchmark SFDA image classification datasets.
\end{itemize}

\section{Related Works}
\label{sec: related works}

\subsection{Unsupervised domain adaptation}

In traditional unsupervised DA, networks are trained jointly on labeled source and unlabeled target dataset to optimize task performance on the target domain~\cite{wilson2020dasurvey}. A popular strategy is to learn domain-invariant features via minimizing domain discrepancy measured by a inter-domain distance or adversarial loss~\cite{Ganin2016DANN,Li2018CDANN,li2018mmd,Sun2016DeepCORAL, zhang2019mdd,hu2020gsda,kang2019can,gu2020rsda,Xu2019sfan}. \cite{lu2020star} learns a distribution of classifiers to better identify local regions of misalignment between source and target domains, and \cite{cui2020gvb,na2021fixbi} facilitate alignment between distant domains by generating intermediate domains.
\cite{Li2021TSA} augments source data to assume target style.
Other methods encourage target representations to be more class-discriminative by learning cluster structure~\cite{tang2020srdc} or minimizing uncertainty~\cite{jin2020mcc}. Amongst confidence-based methods, \cite{gu2020rsda} trains a spherical neural network to select target samples for pseudolabeling, \cite{na2021fixbi,french2018se} selects high confidence predictions as positive pseudolabels and \cite{french2018se} applies mean teacher from semi-supervised learning. Due to the assumed access to both source and target data, these methods are unsuitable when source data is not available for joint training.

\subsection{Source-free domain adaptation}
\label{subsec: source-free domain adaptation}

In SFDA, the source model is adapted with unlabeled target dataset. \cite{kundu2022concurrent,kundu2022balancing} train on source domain with auxiliary tasks or augmentations and \cite{roy2022usfan} calibrates uncertainties with source data to obtain improved source models, but these strategies cannot be applied on the client-side where source data is not accessible.
Some methods make use of local clustering structures in the target dataset to learn class-discriminative features~\cite{yang2021gsfda,Yang2021ExploitingTI,Yang2022AttractingAD}, and learn semantics through self-supervised tasks such as rotation prediction~\cite{xia2021a2net,liang2021shotplus}. \cite{sanqing2022BMD} proposes multi-centric clustering, but the cluster number is not straightforward to select for each dataset. \cite{xia2021a2net} learns a new target-compatible classifier, and \cite{li20203cgan} generates target-style data with a generative adversarial network to improve predictions.
Other methods generates source-like representations~\cite{qiu2021prototypegen} or estimates source feature distribution to align source and target domains~\cite{ding2022sfdade}. \cite{dong2021caida} and \cite{jin2023seprepnet} leverage multiple source domains. \cite{liang2020shot,Kundu2020TowardsIM,li20203cgan} use the source model to generate pseudolabels for the target dataset, and align target samples to the source hypothesis through entropy minimization and information maximization. To improve pseudolabel quality, \cite{kim2021sfda,chen2021noisy,liang2021shotplus} select target samples with low entropy or loss for pseudolabeling and finetuning the source model.
We find that source model outputs may not reliably reflect target pseudolabel accuracy due to domain shift in Figure~\ref{fig: visda_prediction}. We instead make use of an ImageNet pre-trained feature extractor to correct the source bias and produce more reliable pseudolabels. 
\section{Role of Pre-trained Networks in SFDA}
\label{sec: role of pre-trained networks}

\subsection{Conventional role}

In the conventional source-free domain adaptation framework in Figure~\ref{fig: overview_setting}, during source training, the source model is initialized with pre-trained weights and then trained on source data with supervised objective. The pre-trained weights are learned from large and diverse datasets such as ImageNet (IN). Transferring from pre-trained networks is a common practice in computer vision to improve the generalization capability of the trained model and to reduce the requirement on the quantity of training data.

\subsection{Considerations on target generalizability}

In SFDA, the goal is to accurately estimate $p(y_t|x_t)$ for target input $x_t \in \mathcal{X}_t$ and output $y_t \in \mathcal{Y}_t$. For a model with feature extractor (FE) $f$ and classifier (CLF) $k$ and normalizing function $\sigma$, the estimated probabilities are $\left[\hat{p}(y|x_t)\right]_{y\in\mathcal{Y}_t} = \sigma\left(k(f(x_t))\right)$. We note that the source model may not be the network that maximizes the accuracy of $\hat{p}(y_t|x_t)$. Large data pre-trained networks, such as the ImageNet (IN) networks used as initializations for source training, may be more compatible with the target domain instead. That is, class-discriminative information useful for the target domain may be lost from the ImageNet network during source training as the source model learns to fit only the source data distribution. We list the mapping from input to feature space, and from feature to output space for the two models below.

\begin{footnotesize}
\begin{empheq}[box=\fbox]{alignat*=5}
    &\textbf{Input} &&\xrightarrow{\makebox[8mm] {\textbf{FE}}} &&\quad \textbf{Feature} &&\xrightarrow{\makebox[8mm] {\textbf{CLF}}} &&\quad \textbf{Output} \\
    &\textit{IN data} &&\xrightarrow{\makebox[8mm] {$f_*$}} &&\quad \textit{IN feature} &&\xrightarrow{\makebox[8mm] {$k_*$}} &&\quad \textit{IN class} \\    
    &\textit{source} &&\xrightarrow{\makebox[8mm] {$f_s$}} &&\quad \textit{source feature} &&\xrightarrow{\makebox[8mm] {$k_s$}} &&\quad \textit{source class}
\end{empheq}
\end{footnotesize}

The ImageNet and source model are optimized for the accuracy of ImageNet estimates $\hat{p}(y_*|x_*)$ and source domain estimates $\hat{p}(y_s|x_s)$, respectively. Their accuracy on target domain estimates $\hat{p}(y_t|x_t)$ depends on:
\begin{enumerate}[noitemsep]
    \item Similarity between training and target inputs (i.e. images);
    \item Robustness of input-to-feature mapping against training versus target input differences (covariate shift);  \item Similarity between training and target outputs (i.e. class labels).
\end{enumerate}
Pre-trained models can be advantageous in terms of the first two criteria because (1) the larger and more diverse pre-training dataset is not source-biased and may better capture the target input distribution, (2) modern state-of-the-art network architectures are designed to learn more robust input-to-feature mappings, enabling better transfer to target tasks. However, since pre-training and target label space differ, the pre-trained classifier $k_*$ needs to be replaced. One advantage of the source model is that it is trained for the target label space, but it may suffer from a lack of generalization to different input distributions. 

\noindent\textbf{Examples of target generalizability of pre-trained networks.}
Firstly, in Table~\ref{tab: comparison_resnet}, an example where target domain class-discriminative information is lost after source training is the Clipart-to-Real World ($C\rightarrow R$) transfer in Office-Home dataset. The target domain is more distant in style to the source domain than to ImageNet, such that oracle target accuracy (computed by fitting classifier head on the feature extractor using fully-labeled target samples) drops from $86.0\%$ to $83.3\%$ after source training. 
We refer readers to Section~\ref{sec: experiments and results} and \ref{sec: further analysis} for experiment details.

\begin{table}[tb]
\centering
\setlength{\tabcolsep}{2pt}
\begin{adjustbox}{max width=\columnwidth}
\begin{tabular}{l*{6}{c}}
\toprule[1pt]\midrule[0.3pt]

\textbf{Pre-trained feat. extr.} & \multicolumn{2}{c}{$\mathbf{C\rightarrow R}$}  & \multicolumn{2}{c}{$\mathbf{A\rightarrow C}$}  &  \multicolumn{2}{c}{$\mathbf{R\rightarrow A}$} \\ \cmidrule(lr){2-3} \cmidrule(lr){4-5} \cmidrule(lr){6-7}
                    & Oracle    & Co-learn  & Oracle    & Co-learn  & Oracle    & Co-learn \\ \midrule
Source ResNet-50        & 83.3      & 74.1      & \textbf{69.5}      & \emph{53.9}      & \emph{81.8}      & 70.5 \\
ImageNet-1k ResNet-50   & \emph{86.0}      & \emph{79.4}      & 65.5      & 51.8      & 81.2      & \emph{71.1} \\
ImageNet-1k ResNet-101  & \textbf{87.0}      & \textbf{80.4}      & \emph{68.0}      & \textbf{54.6}      & \textbf{82.5}      & \textbf{72.4} \\
\midrule[0.3pt]\bottomrule[1pt]
\end{tabular}
\end{adjustbox}
\vspace{-2mm}
\caption{Oracle target accuracy of pre-trained feature extractor, and classification accuracy of adapted model co-learned with pre-trained feature extractor, on Office-Home. \label{tab: comparison_resnet}}
\vspace{-6mm}
\end{table}
\begin{figure}
  \centering
  \begin{subfigure}{0.48\linewidth}
    \includegraphics[width=\linewidth]{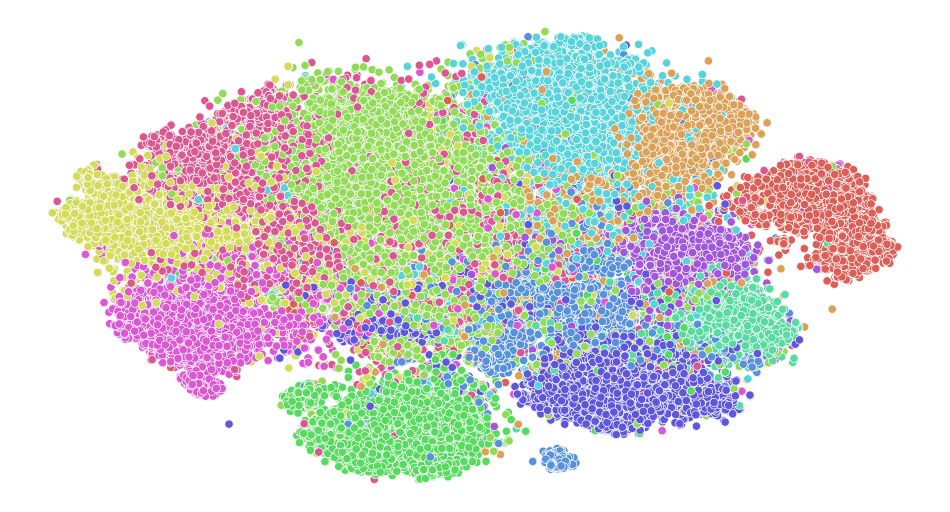}
    \caption{ImageNet-1k ResNet-101}
  \end{subfigure}
  \hfill  
  \begin{subfigure}{0.48\linewidth}
    \includegraphics[width=\linewidth]{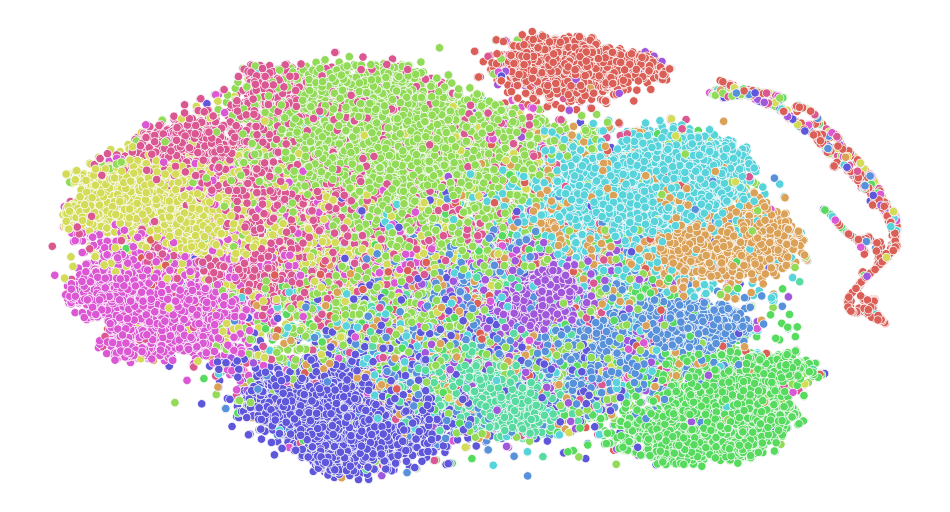}
    \caption{Source ResNet-101}
  \end{subfigure}

  \begin{subfigure}{0.48\linewidth}
    \includegraphics[width=\linewidth]{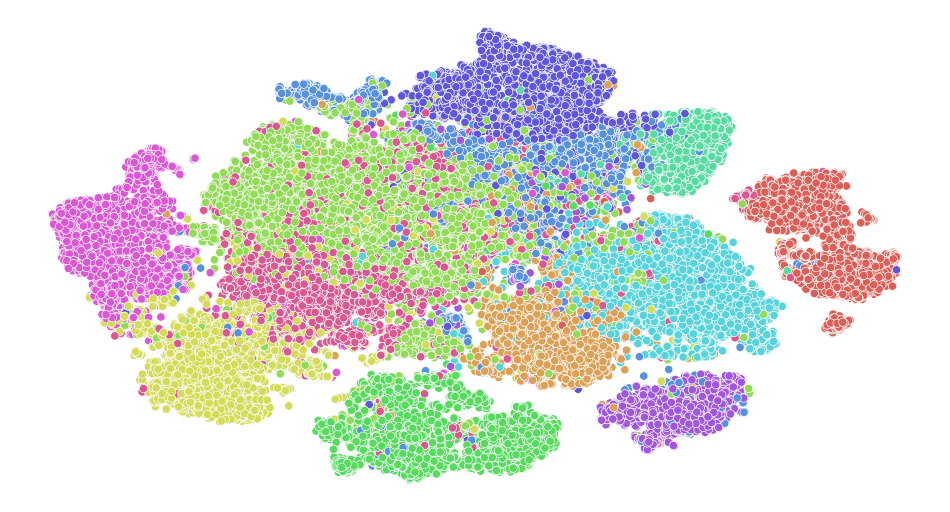}
    \caption{ImageNet-1k Swin-B}
  \end{subfigure}
  \hfill    
  \begin{subfigure}{0.48\linewidth}
    \includegraphics[width=\linewidth]{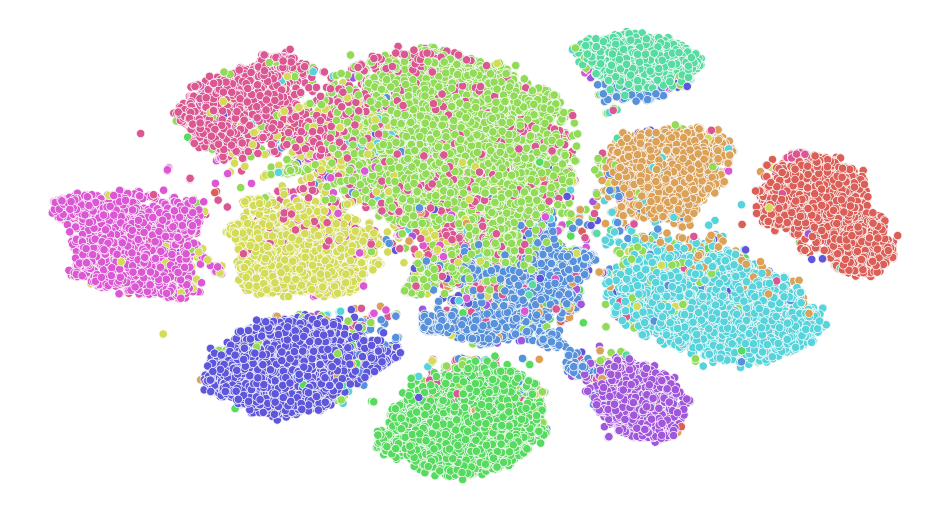}
    \caption{Adapted ResNet-101}
  \end{subfigure}

  \begin{subfigure}{0.8\linewidth}
    \includegraphics[width=\linewidth]{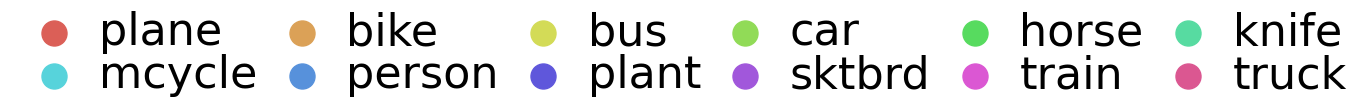}
  \end{subfigure}    
  
  \vspace{-2mm}
  \caption{t-SNE visualization of VisDA-C target domain features by ImageNet-1k ResNet-101, source ResNet-101, ImageNet-1k Swin-B and source ResNet-101 adapted by proposed co-learning with ImageNet-1k Swin-B. Features are extracted at the last pooling layer before the classification head. Samples are colored by class.}
  \label{fig: tsne}
  \vspace{-4mm}
\end{figure}

Secondly, the choice of pre-trained model can improve the robustness of input-to-feature mapping against covariate shift. In Table~\ref{tab: comparison_resnet} for Office-Home dataset, ImageNet ResNet-101 has higher oracle target accuracy than ImageNet ResNet-50. Figure~\ref{fig: tsne} visually compares VisDA-C target domain features extracted at the last pooling layer before the classification head of several models. Swin~\cite{liu2021Swin} is a recent transformer architecture with strong representation learning ability. Even without training on VisDA-C images, the ImageNet Swin-B feature extractor produces more class-discriminative target representations than both ImageNet and source ResNet-101 feature extractors.
\section{Proposed Strategy}
\label{sec: proposed strategy}

From our observations in Section~\ref{sec: role of pre-trained networks}, we propose to distil effective target domain information in pre-trained networks to generate improved pseudolabels to finetune the source model during target adaptation. 

\subsection{Preliminaries}
\label{subsec: preliminaries}

We denote the source and target distributions as 
$\mathcal{P}_s$ and $\mathcal{P}_t$, and observations as $\mathcal{D}_s = \{(x_s^n, y_s^n)\}_{n=1}^{N_s}$ and $\mathcal{D}_t = \{(x_t^n, y_t^n)\}_{n=1}^{N_t}$, for image $x\in\mathcal{X}$ and one-hot classification label $y\in\mathcal{Y}$. The two domains share the same label space $\mathcal{Y}_s = \mathcal{Y}_t$ with $|\mathcal{Y}|=L$ classes, but have different data space $\mathcal{X}_s \neq \mathcal{X}_t$. In SFDA setting, source data $\mathcal{D}_s$ and target labels $\{y_t^n\}_{n=1}^{N_t}$ are not accessible during adaptation. Knowledge on source data is captured by a source model.

The source model trained on $\mathcal{D}_s$ is composed of feature extractor $f_s$ parameterized by $\Theta_s$ that yields learned representations $z_s=f_s(x;\Theta_s)$, and classifier $k_s$ parameterized by $\Psi_s$ that yields logits $g_s=k_s(z_s;\Psi_s)$. Estimated class probabilities are obtained by $p_s=\sigma(g_s)$ for softmax function $\sigma$ where $p_s[i] = \frac{\exp(g_s[i])}{\sum_{j=1}^L \exp{(g_s[j])}}$ for class $i$, and the predicted class is $ \hat{y}_s = \arg\max_i p_s[i]$. 

For a hypothesis $h\in\mathcal{H}$, we refer to $\epsilon_t(h, \ell_t) = E_{x\sim P_t} \epsilon(h(x), \ell_t(x))$ as the target risk of $h$ with respect to the true target labeling function $\ell_t$. To understand the relationship of $h$ and $l_t$ with the source model, we assume an error function $\epsilon$ such as $\epsilon(v,v')=|v-v'|^\alpha$ that satisfies triangle equality following \cite{BenDavid2010domainadaptation, blitzer2007bounds}, then
\begin{equation}
    \epsilon_t(h, \ell_t) \leq \epsilon_t(h, h_p) + \epsilon_t(h_p, \ell_t) \label{eqn: bound}
\end{equation}
where $h_p$ is a pseudolabeling function. The second term $\epsilon_t(h_p, \ell_t)$ is target pseudolabeling error by $h_p$, and the first term is minimized by $h=h_p$. Motivated by the bound in Equation~\ref{eqn: bound}, we propose an iterative strategy to improve the pseudolabeling function $h_p$ and to finetune $h$ towards $h_p$. Our pseudolabeling function leverages a pre-trained network, which is discarded after target adaptation.

\subsection{Co-learning with pre-trained networks}
\label{subsec: proposed sfda}

\begin{table}

\centering
\begin{adjustbox}{max width=0.9\columnwidth}
\begin{tabular}{*{4}{c}}
\toprule[1pt]\midrule[0.3pt]
$\mathbf{\hat{y}_a = \hat{y}_*}$    & $\mathbf{Conf(\hat{y}_a)} > \gamma$    & $\mathbf{Conf(\hat{y}_*)}>\gamma$  & \textbf{Pseudolabel} $\mathbf{\tilde{y}}$ \\ \midrule
\checkmark  & \checkmark / \xmark   & \checkmark / \xmark   & $\hat{y}_a$ \\
\xmark      & \checkmark            & \checkmark            & - \\
\xmark      & \checkmark            & \xmark                & $\hat{y}_a$ \\
\xmark      & \xmark                & \checkmark            & $\hat{y}_*$ \\
\xmark      & \xmark                & \xmark                & - \\
\midrule[0.3pt]\bottomrule[1pt]
\end{tabular}
\end{adjustbox}
\vspace{-2mm}
\caption{MatchOrConf pseudolabeling scheme with adaptation and pre-trained model predictions, $\hat{y}_a$ and $\hat{y}_*$, and confidence threshold $\gamma$. Dash (-) means no pseudolabel. \label{tab: pseudolabel_strategy}}
\vspace{-4mm}
\end{table}
We propose a co-learning strategy to progressively adapt the source model $\{f_s,k_s\}$ with the pre-trained feature extractor $f_*$. The framework consists of two branches: (1) adaptation model branch $\{f_a,k_a\}$ initialized by source model $\{f_s,k_s\}$, (2) pre-trained model branch initialized by $f_*$ and a newly-estimated task classifier $q_*$. Inspired by the nearest-centroid-classifier (NCC) in \cite{liang2020shot}, we construct $q_*$ as a weighted nearest-centroid-classifier where the centroid $\mu_i$ for class $i$ is the sum of $f_*$ features, weighted by estimated class probabilities of the adaptation model:
\begin{align}
    & \mu_i = \frac{\sum_{x} \sigma(k_a(f_a(x;\Theta_a);\Psi_a))[i] f_*(x;\Theta_*) / \|f_*(x;\Theta_*)\|}{\sum_{x} \sigma(k_a(f_a(x;\Theta_a);\Psi_a))[i]} \label{eqn: pretrained_centroid} \\
    & g_*[i] = q_*(f_*(x;\Theta_*))[i] = \frac{f_*(x;\Theta_*) \cdot \mu_i}{\|f_*(x;\Theta_*)\| \|\mu_i\|} \\
    & p_* = \sigma(g_* / Temperature) \label{eqn: pretrained_prob}
\end{align}
The weighted NCC leverages the target domain class-discriminative cluster structures in $f_*$ features. In Equation~\ref{eqn: pretrained_prob}, predictions $p_*$ are sharpened with $Temperature = 0.01$ since the logits $g_*$ calculated through cosine similarity are bounded in $[-1,1]$

\begin{algorithm}[tb]
\caption{Proposed co-learning strategy \label{alg}}

\textbf{Input:} Target images $\{x_t^n\}_{n=1}^{N_t}$;
source model $k_s \circ f_s$ parameterized by $(\Theta_s, \Psi_s)$; pre-trained feature extractor $f_*$ parameterized by $\Theta_*$; confidence threshold $\gamma$; learning rate $\eta$; \# episodes $I$;
\begin{algorithmic}[1]

\Procedure{Co-learning}{}
\State Initialize weights of adaption model branch $k_a \circ f_a$ by $(\Theta_a, \Psi_a) \leftarrow (\Theta_s, \Psi_s)$
\State Initialize pre-trained model branch classifier $q_*$ by computing centroids according to Equation~\ref{eqn: pretrained_centroid}

\For{episode $i=1:I$}
\State Construct pseudolabel dataset $\Tilde{\mathcal{D}}_t$ by MatchOrConf scheme with threshold $\gamma$ in Table~\ref{tab: pseudolabel_strategy}
\State Compute loss $L_{co-learn}(\Theta_a)$ in Equation~\ref{eqn: colearn}
\State Update weights $\Theta_a \leftarrow \Theta_a - \eta \nabla L_{co-learn}(\Theta_a)$
\State Update centroids in $q_*$ according to Equation~\ref{eqn: pretrained_centroid}
\EndFor
\EndProcedure

\State \Return Adaptation model $k_a \circ f_a$ with weights $(\Theta_a, \Psi_a)$
\end{algorithmic}
\end{algorithm}

\noindent\textbf{Co-learning:}
We alternate updates on the adaptation and pre-trained model branch as each adapts to the target domain and produces more accurate predictions to improve the other branch. Each round of update on the two branches is referred to as a co-learning episode. For the adaptation model branch, we freeze classifier $k_a$ and update feature extractor $f_a$ to correct its initial bias towards the source domain. We finetune $f_a$ with cross-entropy loss on improved pseudolabeled samples, further described in the next paragraph. For the pre-trained model branch, we freeze feature extractor $f_*$ to retain the target-compatible features and update classifier $q_*$. Since $f_*$ is not updated throughout the adaptation process, only a single pass of target data through $f_*$ is needed to construct a feature bank; previous works similarly required memory banks of source model features~\cite{Yang2022AttractingAD,yang2021gsfda,Yang2021ExploitingTI}. The centroids are updated by Equation~\ref{eqn: pretrained_centroid} using estimated class probabilities $\sigma(k_a(f_a(x;\Theta_a);\Psi_a))$ of the current adaptation model. Improvements in the adaptation model predictions each episode give more accurate class centroids in the pre-trained model branch. 

\noindent\textbf{Improved pseudolabels:}
We improve pseudolabels each episode by combining predictions from the two branches using the MatchOrConf scheme in Table~\ref{tab: pseudolabel_strategy}. Denoting $\hat{y}_a$ and $\hat{y}_*$ as the adaptation and pre-trained model predictions, respectively, the pseudolabel $\Tilde{y}$ given is $\hat{y}_a (=\hat{y}_*)$ if the predicted classes match, and the predicted class with higher confidence level otherwise. Confidence level is determined by a threshold $\gamma$. Remaining samples are not pseudolabeled. The co-learning objective for the adaptation model is
\begin{equation}
    L_{co-learn}(\Theta_a) = -\sum_{(x,\Tilde{y})\in \Tilde{D}_t} \Tilde{y} \cdot \log( p_a ) \label{eqn: colearn}
\end{equation}
where $p_a = \sigma(k_a( f_a(x; \Theta_a) ; \Psi_a))$ and $\Tilde{\mathcal{D}}_t$ is pseudolabeled target dataset. Algorithm~\ref{alg} summarizes the strategy. We can incorporate the strategy into existing SFDA methods by replacing the pseudolabels used with our co-learned pseudolabels or by adding $L_{co-learn}$ to the learning objective. 
\section{Experiments and Results}
\label{sec: experiments and results}

We evaluate on 4 benchmark image classification datasets for domain adaptation. We describe experimental setups in Section~\ref{subsec: experimental setups} and results in Section~\ref{subsec: results}.
\begin{table*}[tb]
\centering
\begin{adjustbox}{max width=0.80\textwidth}
\begin{tabular}{l*{7}{c}l}
\toprule[1pt]\midrule[0.3pt]
\textbf{Method}                 & \textbf{SF} & $\mathbf{A\rightarrow D}$ & $\mathbf{A\rightarrow W}$ & $\mathbf{D\rightarrow A}$  
                                & $\mathbf{D\rightarrow W}$ & $\mathbf{W\rightarrow A}$ & $\mathbf{W\rightarrow D}$ &\textbf{Avg} \\ \midrule                                
MDD \cite{zhang2019mdd}         & \xmark & 93.5  & 94.5  & 74.6  & 98.4  & 72.2  & \underline{\textbf{100.0}} & 88.9 \\
GVB-GD \cite{cui2020gvb}        & \xmark & 95.0  & 94.8  & 73.4  & 98.7  & 73.7  & \underline{\textbf{100.0}} & 89.3 \\
MCC \cite{jin2020mcc}           & \xmark & 95.6  & 95.4  & 72.6  & 98.6  & 73.9  & \underline{\textbf{100.0}} & 89.4 \\
GSDA \cite{hu2020gsda}          & \xmark & 94.8  & \textbf{95.7}  & 73.5  & 99.1  & 74.9  & \underline{\textbf{100.0}} & 89.7 \\
CAN \cite{kang2019can}          & \xmark & 95.0  & 94.5  & \textbf{78.0}  & 99.1  & 77.0  & 99.8  & 90.6 \\
SRDC \cite{tang2020srdc}        & \xmark & \textbf{95.8}  & \textbf{95.7}  & 76.7  & \textbf{99.2}  & \textbf{77.1}  & \underline{\textbf{100.0}} & \textbf{90.8} \\ \midrule
Source Only                     & \cmark & 81.9  & 78.0  & 59.4  & 93.6  & 63.4  & 98.8  & 79.2 \\
Co-learn (w/ ResNet-50)         & \cmark & 93.6  & 90.2  & 75.7  & 98.2  & 72.5  & 99.4  & 88.3 \\
Co-learn (w/ Swin-B)            & \cmark & \textbf{97.4}  & \textbf{98.2}  & \underline{\textbf{84.5}}  & \textbf{99.1}  & \underline{\textbf{82.2}}  & \underline{\textbf{100.0}} & \underline{\textbf{93.6}} \\ \hdashline
SHOT$^\dagger$ \cite{liang2020shot}         & \cmark & 95.0 & 90.4 & 75.2 & \textbf{98.9} & 72.8 & 99.8 & 88.7 \\
\quad w/ Co-learn (w/ ResNet-50)            &        & 94.2 & 90.2 & 75.7 & 98.2 & 74.4 & \underline{\textbf{100.0}} & 88.8 $\green{(\uparrow 0.1)}$ \\
\quad w/ Co-learn (w/ Swin-B)               &        & \textbf{95.8} & \textbf{95.6} & \textbf{78.5} & \textbf{98.9} & \textbf{76.7} & 99.8 & \textbf{90.9} $\green{(\uparrow 2.2)}$ \\ \hdashline
SHOT++$^\dagger$ \cite{liang2021shotplus}   & \cmark & 95.6 & 90.8 & 76.0 & 98.2 & 74.6 & \underline{\textbf{100.0}} & 89.2 \\
\quad w/ Co-learn (w/ ResNet-50)            &        & 95.0 & 90.7 & 76.4 & 97.7 & 74.9 & 99.8 & 89.1 $\red{(\downarrow 0.1)}$ \\
\quad w/ Co-learn (w/ Swin-B)               &        & \textbf{96.6} & \textbf{93.8} & \textbf{79.8} & \textbf{98.9} & \textbf{78.0} & \underline{\textbf{100.0}} & \textbf{91.2} $\green{(\uparrow 2.0)}$ \\ \hdashline
NRC$^\dagger$ \cite{Yang2021ExploitingTI}   & \cmark & 92.0 & 91.6 & 74.5 & 97.9 & 74.8 & \underline{\textbf{100.0}} & 88.5 \\
\quad w/ Co-learn (w/ ResNet-50)            &        & 95.4 & 89.9 & 76.5 & 98.0 & 76.0 & 99.8 & 89.3 $\green{(\uparrow 0.8)}$\\
\quad w/ Co-learn (w/ Swin-B)               &        & \textbf{96.2} & \textbf{94.3} & \textbf{79.1} & \textbf{98.7} & \textbf{78.5} & \underline{\textbf{100.0}} & \textbf{91.1} $\green{(\uparrow 2.6)}$\\ \hdashline
AaD$^\dagger$ \cite{Yang2022AttractingAD}   & \cmark & 94.4 & 93.3 & 75.9 & 98.4 & 76.3 & 99.8 & 89.7 \\
\quad w/ Co-learn (w/ ResNet-50)            &        & 96.6 & 92.5 & 77.3 & 98.9 & 76.6 & 99.8 & 90.3 $\green{(\uparrow 0.6)}$\\
\quad w/ Co-learn (w/ Swin-B)               &        & \underline{\textbf{97.6}} & \underline{\textbf{98.7}} & \textbf{82.1} & \underline{\textbf{99.3}} & \textbf{80.1} & \underline{\textbf{100.0}} & \textbf{93.0} $\green{(\uparrow 3.3)}$\\
\midrule[0.3pt]\bottomrule[1pt]
\end{tabular}
\end{adjustbox}
\vspace{-2mm}
\caption{Office-31: 31-class classification accuracy of adapted ResNet-50. For proposed strategy, ImageNet-1k feature extractor used is given in parenthesis. SF denotes source-free. $\dagger$ denotes reproduced results. \label{tab: office31_results}}
\vspace{-2mm}
\end{table*}
\begin{table*}[tb]
\centering
 \setlength{\tabcolsep}{2pt}
\begin{adjustbox}{max width=\textwidth}
\begin{tabular}{l*{13}{c}l}
\toprule[1pt]\midrule[0.3pt]
\textbf{Method}                 & \textbf{SF} & $\mathbf{A\rightarrow C}$ & $\mathbf{A\rightarrow P}$ & $\mathbf{A\rightarrow R}$  
                                & $\mathbf{C\rightarrow A}$ & $\mathbf{C\rightarrow P}$ & $\mathbf{C\rightarrow R}$ 
                                & $\mathbf{P\rightarrow A}$ & $\mathbf{P\rightarrow C}$ & $\mathbf{P\rightarrow R}$
                                & $\mathbf{R\rightarrow A}$ & $\mathbf{R\rightarrow C}$ & $\mathbf{R\rightarrow P}$ & \textbf{Avg}\\ \midrule
GSDA \cite{hu2020gsda}          & \xmark & \textbf{61.3}  & 76.1  & 79.4  & 65.4  & 73.3  & 74.3  & 65.0  & 53.2  & 80.0  & 72.2  & 60.6  & 83.1  & 70.3 \\
GVB-GD \cite{cui2020gvb}        & \xmark & 57.0  & 74.7  & 79.8  & 64.6  & 74.1  & 74.6  & 65.2  & 55.1  & 81.0  & 74.6  & 59.7  & 84.3  & 70.4 \\
RSDA \cite{gu2020rsda}          & \xmark & 53.2  & \textbf{77.7}  & \textbf{81.3}  & 66.4  & 74.0  & 76.5  & 67.9  & 53.0  & \textbf{82.0}  & 75.8  & 57.8  & 85.4  & 70.9 \\
TSA \cite{Li2021TSA}            & \xmark & 57.6  & 75.8  & 80.7  & 64.3  & 76.3  & 75.1  & 66.7  & 55.7  & 81.2  & 75.7  & 61.9  & 83.8  & 71.2 \\
SRDC \cite{tang2020srdc}        & \xmark & 52.3  & 76.3  & 81.0  & \textbf{69.5}  & 76.2  & 78.0  & \textbf{68.7}  & 53.8  & 81.7  & 76.3  & 57.1  & 85.0  & 71.3 \\
FixBi \cite{na2021fixbi}        & \xmark & 58.1  & 77.3  & 80.4  & 67.7  & \textbf{79.5}  & \textbf{78.1}  & 65.8  & \textbf{57.9}  & 81.7  & \textbf{76.4}  & \textbf{62.9}  & \textbf{86.7}  & \textbf{72.7} \\ \midrule
Source Only                     & \cmark & 43.5  & 67.1  & 74.2  & 51.5  & 62.2  & 63.3  & 51.4  & 40.7  & 73.2  & 64.6  & 45.8  & 77.6  & 59.6 \\
Co-learn (w/ ResNet-50)         & \cmark & 51.8  & 78.9  & 81.3  & 66.7  & 78.8  & 79.4  & 66.3  & 50.0  & 80.6  & 71.1  & 53.7  & 81.3  & 70.0\\
Co-learn (w/ Swin-B)            & \cmark & \underline{\textbf{69.6}}  & \underline{\textbf{89.5}}  & \underline{\textbf{91.2}}  & \underline{\textbf{82.7}}  & \underline{\textbf{88.4}}  & \underline{\textbf{91.3}}  & \underline{\textbf{82.6}}  & \textbf{68.5}  & \underline{\textbf{91.5}}  & \underline{\textbf{82.8}}  & \underline{\textbf{71.3}}  & \underline{\textbf{92.1}}  & \underline{\textbf{83.5}} \\ \hdashline
SHOT$^\dagger$ \cite{liang2020shot}         & \cmark & 55.8 & 79.6 & 82.0 & 67.4 & 77.9 & 77.9 & 67.6 & 55.6 & 81.9 & 73.3 & 59.5 & 84.0 & 71.9\\
\quad w/ Co-learn (w/ ResNet-50)            &        & 56.3 & 79.9 & 82.9 & 68.5 & 79.6 & 78.7 & 68.1 & 54.8 & 82.5 & 74.5 & 59.0 & 83.6 & 72.4 $\green{(\uparrow 0.5)}$ \\
\quad w/ Co-learn (w/ Swin-B)               &        & \textbf{61.7} & \textbf{82.9} & \textbf{85.3} & \textbf{72.7} & \textbf{80.5} & \textbf{82.0} & \textbf{71.6} & \textbf{60.4} & \textbf{84.5} & \textbf{76.0} & \textbf{64.3} & \textbf{86.7} & \textbf{75.7} $\green{(\uparrow 3.8)}$ \\ \hdashline
SHOT++$^\dagger$ \cite{liang2021shotplus}   & \cmark & 57.1 & 79.5 & 82.6 & 68.5 & 79.5 & 78.6 & 68.3 & 56.1 & 82.9 & 74.0 & 59.8 & 85.0 & 72.7 \\
\quad w/ Co-learn (w/ ResNet-50)            &        & 57.7 & 81.1 & 84.0 & 69.2 & 79.8 & 79.2 & 69.1 & 57.7 & 82.9 & 73.7 & 60.1 & 85.0 & 73.3 $\green{(\uparrow 0.6)}$ \\
\quad w/ Co-learn (w/ Swin-B)               &        & \textbf{63.7} & \textbf{83.0} & \textbf{85.7} & \textbf{72.6} & \textbf{81.5} & \textbf{83.8} & \textbf{72.0} & \textbf{59.9} & \textbf{85.3} & \textbf{76.3} & \textbf{65.3} & \textbf{86.6} & \textbf{76.3} $\green{(\uparrow 3.6)}$ \\ \hdashline
NRC$^\dagger$ \cite{Yang2021ExploitingTI}   & \cmark & 58.0 & 79.3 & 81.8 & 70.1 & 78.7 & 78.7 & 63.5 & 57.0 & 82.8 & 71.6 & 58.2 & 84.3 & 72.0 \\
\quad w/ Co-learn (w/ ResNet-50)            &        & 56.1 & 80.3 & 83.0 & 70.3 & 81.3 & 80.9 & 67.7 & 53.9 & 83.7 & 72.5 & 57.9 & 83.4 & 72.6 $\green{(\uparrow 0.6)}$ \\
\quad w/ Co-learn (w/ Swin-B)               &        & \textbf{67.8} & \textbf{86.4} & \textbf{89.1} & \textbf{80.7} & \textbf{87.5} & \textbf{89.3} & \textbf{77.8} & \underline{\textbf{68.8}} & \textbf{89.7} & \textbf{81.6} & \textbf{68.7} & \textbf{89.9} & \textbf{81.4} $\green{(\uparrow 9.4)}$ \\ \hdashline
AaD$^\dagger$ \cite{Yang2022AttractingAD}   & \cmark & 58.7 & 79.8 & 81.4 & 67.5 & 79.4 & 78.7 & 64.7 & 56.8 & 82.5 & 70.3 & 58.0 & 83.3 & 71.8 \\
\quad w/ Co-learn (w/ ResNet-50)            &        & 57.7 & 80.4 & 83.3 & 70.1 & 80.1 & 80.6 & 66.6 & 55.5 & 84.1 & 72.1 & 57.6 & 84.3 & 72.7 $\green{(\uparrow 0.9)}$ \\
\quad w/ Co-learn (w/ Swin-B)               &        & \textbf{65.1} & \textbf{86.0} & \textbf{87.0} & \textbf{76.8} & \textbf{86.3} & \textbf{86.5} & \textbf{74.4} & \textbf{66.1} & \textbf{87.7} & \textbf{77.9} & \textbf{66.1} & \textbf{88.4} & \textbf{79.0} $\green{(\uparrow 7.2)}$\\
\midrule[0.3pt]\bottomrule[1pt]
\end{tabular}
\end{adjustbox}
\vspace{-2mm}
\caption{Office-Home: 65-class classification accuracy of adapted ResNet-50. For proposed strategy, ImageNet-1k feature extractor used is given in parenthesis. SF denotes source-free. $\dagger$ denotes reproduced results. \label{tab: officehome_results}}
\vspace{-4mm}
\end{table*}
\begin{table*}[tb]
\centering
 \setlength{\tabcolsep}{2pt}
\begin{adjustbox}{max width=0.95\textwidth}
\begin{tabular}{l*{13}{c}l}
\toprule[1pt]\midrule[0.3pt]

\textbf{Method}                 & \textbf{SF} & $\mathbf{C\rightarrow P}$ & $\mathbf{C\rightarrow R}$ & $\mathbf{C\rightarrow S}$  
                                & $\mathbf{P\rightarrow C}$ & $\mathbf{P\rightarrow R}$ & $\mathbf{P\rightarrow S}$ 
                                & $\mathbf{R\rightarrow C}$ & $\mathbf{R\rightarrow P}$ & $\mathbf{R\rightarrow S}$
                                & $\mathbf{S\rightarrow C}$ & $\mathbf{S\rightarrow P}$ & $\mathbf{S\rightarrow R}$ & \textbf{Avg}\\ \midrule
Source Only                     & \cmark & 47.2 & 61.7 & 50.7 & 47.2 & 71.7 & 44.3 & 58.2 & 63.2 & 49.2 & 59.5 & 52.4 & 60.5 & 55.5 \\
Co-learn (w/ ResNet-50)         & \cmark & 58.7 & 75.7 & 51.9 & 54.9 & 76.6 & 46.1 & 61.4 & 63.7 & 49.1 & 65.0 & 62.7 & 76.7 & 61.9 \\
Co-learn (w/ Swin-B)            & \cmark & \underline{\textbf{69.1}} & \underline{\textbf{85.0}} & \textbf{61.3} & \textbf{68.7} & \underline{\textbf{87.3}} & \textbf{60.6} & \textbf{70.3} & \textbf{71.5} & \textbf{59.5} & \textbf{70.1} & \underline{\textbf{72.9}} & \underline{\textbf{85.2}} & \textbf{71.8} \\ \hdashline
SHOT$^\dagger$ \cite{liang2020shot}         & \cmark & 62.0 & 78.0 & 59.9 & 63.9 & 78.8 & 57.4 & 68.7 & 67.8 & 57.7 & 71.5 & 65.5 & 76.3 & 67.3 \\
\quad w/ Co-learn (w/ ResNet-50)            &        & 62.4 & 78.0 & 60.1 & 63.8 & 79.2 & 57.3 & 68.0 & 67.6 & 58.2 & 71.5 & 65.7 & 75.9 & 67.3 $\green{(=)}$ \\
\quad w/ Co-learn (w/ Swin-B)               &        & \textbf{64.8} & \textbf{82.3} & \textbf{63.1} & \textbf{68.9} & \textbf{84.0} & \textbf{62.7} & \textbf{72.3} & \textbf{70.6} & \textbf{61.7} & \textbf{74.0} & \textbf{69.2} & \textbf{83.6} & \textbf{71.4} $\green{(\uparrow 4.1)}$ \\ \hdashline
SHOT++$^\dagger$ \cite{liang2021shotplus}   & \cmark & 63.0 & 80.4 & 62.5 & 66.9 & 80.0 & 60.0 & 71.2 & 68.7 & 61.2 & 72.8 & 66.6 & 78.1 & 69.3 \\
\quad w/ Co-learn (w/ ResNet-50)            &        & 63.0 & 80.0 & 61.9 & 65.9 & 80.2 & 60.6 & 70.6 & 68.8 & 61.2 & 73.3 & 67.2 & 78.2 & 69.2 $\red{(\downarrow 0.1)}$ \\
\quad w/ Co-learn (w/ Swin-B)               &        & \textbf{65.9} & \textbf{82.7} & \underline{\textbf{65.3}} & \underline{\textbf{71.0}} & \textbf{84.0} & \textbf{64.9} & \textbf{73.5} & \textbf{71.7} & \underline{\textbf{64.6}} & \textbf{75.7} & \textbf{70.6} & \textbf{84.5} & \underline{\textbf{72.9}} $\green{(\uparrow 3.6)}$ \\ \hdashline
AaD$^\dagger$ \cite{Yang2022AttractingAD}   & \cmark & 62.5 & 78.7 & 59.4 & 65.3 & 79.9 & 61.3 & 69.3 & 68.6 & 57.1 & 72.9 & 67.5 & 77.4 & 68.3 \\
\quad w/ Co-learn (w/ ResNet-50)            &        & 63.5 & 79.7 & 62.1 & 66.4 & 81.5 & 63.0 & 72.3 & 69.7 & 60.7 & 74.5 & 69.3 & 79.8 & 70.2 $\green{(\uparrow 1.9)}$ \\
\quad w/ Co-learn (w/ Swin-B)               &        & \textbf{66.8} & \textbf{81.0} & \textbf{63.8} & \textbf{70.4} & \textbf{84.0} & \underline{\textbf{65.4}} & \underline{\textbf{74.6}} & \underline{\textbf{72.1}} & \textbf{63.8} & \underline{\textbf{76.4}} & \textbf{71.2} & \textbf{82.8} & \textbf{72.7} $\green{(\uparrow 4.4)}$\\
\midrule[0.3pt]\bottomrule[1pt]
\end{tabular}
\end{adjustbox}
\vspace{-2mm}
\caption{DomainNet-126: 126-class classification accuracy of adapted ResNet-50. For proposed strategy, ImageNet-1k feature extractor used is given in parenthesis. SF denotes source-free. $\dagger$ denotes reproduced results. \label{tab: domainnet_results}}
\vspace{-2mm}
\end{table*}
\begin{table*}[tb]
\centering
\setlength{\tabcolsep}{3pt}
\begin{adjustbox}{max width=0.92\textwidth}
\begin{tabular}{l*{13}{c}l}
\toprule[1pt]\midrule[0.3pt]
\textbf{Method}                 & \textbf{SF} & \textbf{plane} & \textbf{bike} & \textbf{bus} & \textbf{car} & \textbf{horse} & \textbf{knife} & \textbf{mcycle} & \textbf{person} & \textbf{plant} & \textbf{sktbrd} & \textbf{train} & \textbf{truck} & \textbf{Avg}\\ \midrule
SFAN \cite{Xu2019sfan}          & \xmark & 93.6  & 61.3  & 84.1  & 70.6  & 94.1  & 79.0  & 91.8  & 79.6  & 89.9  & 55.6  & 89.0  & 24.4  & 76.1 \\
MCC \cite{jin2020mcc}           & \xmark & 88.7  & 80.3  & 80.5  & 71.5  & 90.1  & 93.2  & 85.0  & 71.6  & 89.4  & 73.8  & 85.0  & 36.9  & 78.8 \\
STAR \cite{lu2020star}          & \xmark & 95.0  & 84.0  & 84.6  & 73.0  & 91.6  & 91.8  & 85.9  & 78.4  & 94.4  & 84.7  & 87.0  & 42.2  & 82.7 \\
SE \cite{french2018se}          & \xmark & 95.9  & 87.4  & 85.2  & 58.6  & 96.2  & 95.7  & 90.6  & 80.0  & 94.8  & 90.8  & 88.4  & 47.9  & 84.3 \\
CAN \cite{kang2019can}          & \xmark & \textbf{97.0}  & 87.2  & 82.5  & 74.3  & \textbf{97.8}  & \textbf{96.2}  & 90.8  & 80.7  & 96.6  & \underline{\textbf{96.3}}  & 87.5  & \textbf{59.9}  & \textbf{87.2} \\
FixBi \cite{na2021fixbi}        & \xmark & 96.1  & \textbf{87.8}  & \underline{\textbf{90.5}}  & \underline{\textbf{90.3}}  & 96.8  & 95.3  & \textbf{92.8}  & \underline{\textbf{88.7}}  & \textbf{97.2}  & 94.2  & \textbf{90.9}  & 25.7  & \textbf{87.2} \\ \midrule
Source Only                     & \cmark & 51.5  & 15.3  & 43.4  & 75.4  & 71.2  & 6.8   & 85.5  & 18.8  & 49.4  & 46.4  & 82.1  & 5.4   & 45.9 \\
Co-learn (w/ ResNet-101)        & \cmark & 96.5  & 78.9  & 77.5  & 75.7  & 94.6  & 95.8  & 89.1  & 77.7  & 90.5  & 91.0  & 86.2  & \textbf{51.5}  & 83.7 \\
Co-learn (w/ Swin-B)            & \cmark & \underline{\textbf{99.0}}  & \textbf{90.0}  & \textbf{84.2}  & \textbf{81.0}  & \underline{\textbf{98.1}}  & \textbf{97.9}  & \underline{\textbf{94.9}}  & \textbf{80.1}  & \textbf{94.8}  & \textbf{95.9}  & \textbf{94.4}  & 48.1  & \textbf{88.2} \\ \hdashline
SHOT$^\dagger$ \cite{liang2020shot}         & \cmark & 95.3 & 87.1 & 79.1 & 55.1 & 93.2 & 95.5 & 79.5 & 79.6 & 91.6 & 89.5 & 87.9 & 56.0 & 82.4 \\
\quad w/ Co-learn (w/ ResNet-101)           &        & 94.9 & 84.8 & 77.7 & \textbf{63.0} & 94.1 & 95.6 & 85.6 & 81.0 & \textbf{93.0} & \textbf{92.2} & 86.4 & 60.4 & 84.1 $\green{(\uparrow 1.7)}$ \\
\quad w/ Co-learn (w/ Swin-B)               &        & \textbf{96.0} & \textbf{88.1} & \textbf{81.0} & \textbf{63.0} & \textbf{94.3} & \textbf{95.9} & \textbf{87.1} & \textbf{81.8} & 92.8 & 91.9 & \textbf{90.1} & \underline{\textbf{60.5}} & \textbf{85.2} $\green{(\uparrow 2.8)}$ \\ \hdashline
SHOT++$^\dagger$ \cite{liang2021shotplus}   & \cmark & 94.5 & 88.5 & \textbf{90.4} & 84.6 & \textbf{97.9} & \underline{\textbf{98.6}} & 91.9 & 81.8 & 96.7 & 91.5 & 93.8 & 31.3 & 86.8 \\
\quad w/ Co-learn (w/ ResNet-101)           &        & 97.9 & 88.6 & 86.8 & \textbf{86.7} & \textbf{97.9} & \underline{\textbf{98.6}} & \textbf{92.4} & 83.6 & 97.4 & 92.5 & 94.4 & 32.5 & 87.4 $\green{(\uparrow 0.6)}$ \\
\quad w/ Co-learn (w/ Swin-B)               &        & \textbf{98.0} & \underline{\textbf{91.1}} & 88.6 & 83.2 & 97.8 & 97.8 & 92.0 & \textbf{85.8} & \underline{\textbf{97.6}} & \textbf{93.2} & \underline{\textbf{95.0}} & \textbf{43.5} & \textbf{88.6} $\green{(\uparrow 1.8)}$ \\ \hdashline
NRC$^\dagger$ \cite{Yang2021ExploitingTI}   & \cmark & 96.8 & \textbf{92.0} & 83.8 & 57.2 & 96.6 & 95.3 & 84.2 & 79.6 & \textbf{94.3} & 93.9 & 90.0 & 59.8 & 85.3  \\
\quad w/ Co-learn (w/ ResNet-101)           &        & 96.9 & 89.2 & 81.1 & 65.5 & 96.3 & 96.1 & \textbf{89.8} & 80.6 & 93.7 & \textbf{95.4} & 88.8 & 60.0 & 86.1 $\green{(\uparrow 0.8)}$\\
\quad w/ Co-learn (w/ Swin-B)               &        & \textbf{97.4} & 91.3 & \textbf{84.5} & \textbf{65.8} & \textbf{96.9} & \textbf{97.6} & 88.8 & \textbf{82.0} & 93.8 & 94.7 & \textbf{91.1} & \textbf{61.6} & \textbf{87.1} $\green{(\uparrow 1.8)}$\\ \hdashline
AaD$^\dagger$ \cite{Yang2022AttractingAD}   & \cmark & 96.9 & \textbf{90.2} & \textbf{85.7} & 82.8 & 97.4 & 96.0 & 89.7 & 83.2 & \textbf{96.8} & 94.4 & 90.8 & 49.0 & 87.7 \\
\quad w/ Co-learn (w/ ResNet-101)           &        & \textbf{97.7} & 87.9 & 84.8 & 79.6 & \textbf{97.6} & \textbf{97.5} & \textbf{92.4} & 83.7 & 95.3 & 94.2 & 90.3 & \textbf{57.4} & 88.2 $\green{(\uparrow 0.5)}$\\
\quad w/ Co-learn (w/ Swin-B)               &        & 97.6 & \textbf{90.2} & 85.0 & \textbf{83.1} & \textbf{97.6} & 97.1 & 92.1 & \textbf{84.9} & \textbf{96.8} & \textbf{95.1} & \textbf{92.2} & 56.8 & \underline{\textbf{89.1}} $\green{(\uparrow 1.4)}$\\
\midrule[0.3pt]\bottomrule[1pt]
\end{tabular}
\end{adjustbox}
\vspace{-2mm}
\caption{VisDA-C: 12-class classification accuracy of adapted ResNet-101. For proposed strategy, ImageNet-1k feature extractor used is given in parenthesis. SF denotes source-free. $\dagger$ denotes reproduced results. \label{tab: visda_results}}
\vspace{-4
mm}
\end{table*}

\subsection{Experimental setups}
\label{subsec: experimental setups}

\noindent\textbf{Datasets.}
\textbf{Office-31}~\cite{Saenko2010AdaptingVC} has 31 categories of office objects in 3 domains: Amazon (A), Webcam (W) and DSLR (D).
\textbf{Office-Home}~\cite{Venkateswara2017DeepHN} has 65 categories of everyday objects in 4 domains: Art (A), Clipart (C), Product (P) and Real World (R).
\textbf{DomainNet}~\cite{peng2019moment} is a challenging dataset with a total of 6 domains and 345 classes. Following \cite{litrico2023uncertainty}, we evaluate on 4 domains: Clipart (C), Painting (P), Real (R) and Sketch (S) with a subset of 126 classes. 
\textbf{VisDA-C}~\cite{Peng2017VisDATV} is a popular 12-class dataset for evaluating synthetic-to-real shift, with synthetic rendering of 3D models in source domain and Microsoft COCO real images in target domain.
We report classification accuracy on the target domain for all domain pairs in Office-31, Office-Home and DomainNet, and average per-class accuracy on the real domain in VisDA-C.

\noindent\textbf{Implementation details.}
We follow the network architecture and training scheme in \cite{liang2020shot,liang2021shotplus} to train source models: Office-31, Office-Home and DomainNet use ResNet-50 and VisDA-C uses ResNet-101 initialized with ImageNet-1k weights for feature extractor plus a 2-layer linear classifier with weight normalization, trained on labeled source data.
For our proposed strategy, we experiment with the following ImageNet-1k feature extractors curated on Torch~\cite{torchmodels} and Hugging Face~\cite{huggingfacemodels}  for co-learning: ResNet-50, ResNet-101, ConvNeXt-S, Swin-S, ConvNeXt-B and Swin-B, where S and B denote the small and base versions of the architectures, respectively.
This list is not exhaustive and is meant as a demonstration that state-of-the-art networks can be successfully plugged into our proposed framework. ConvNeXt convolutional networks~\cite{liu2022convnet} and Swin transformers~\cite{liu2021Swin} are recently-released architectures for computer vision tasks that demonstrated improved robustness to domain shifts~\cite{kim2022broad}. During target adaptation, we train using SGD optimizer for 15 episodes, with batch size 50 and learning rate 0.01 decayed to 0.001 after 10 episodes. We set confidence threshold $\gamma=0.5$ for Office-31, Office-Home and DomainNet and $\gamma=0.1$ for VisDA-C, with further analysis in Section~\ref{sec: further analysis}. We also apply our strategy on existing methods. For SHOT~\cite{liang2020shot} and SHOT++~\cite{liang2021shotplus} with pseudolabeling components, we replace the pseudolabels used with co-learned pseudolabels. For NRC~\cite{Yang2021ExploitingTI} and AaD~\cite{Yang2022AttractingAD} originally without pseudolabeling components, we add $0.3 L_{co-learn}$ to the training objective where the coefficient $0.3$ follows that in SHOT~\cite{liang2020shot} and SHOT++~\cite{liang2021shotplus}.

\subsection{Results}
\label{subsec: results}

We report results for co-learning with the ImageNet-1k network (ResNet-50 or ResNet-101) used for source model initialization and Swin-B in Table~\ref{tab: office31_results}, \ref{tab: officehome_results}, \ref{tab: domainnet_results} and \ref{tab: visda_results}. Best performance in each set of comparison is \textbf{bolded}, and best performance for each source-target transfer is \underline{underlined}. Full results on other architectures are in Appendix .

\noindent\textbf{Reusing pre-trained network from source model initialization:} Reusing the same ImageNet network from source model initialization for co-learning can improve the original source model performance. Since Office domains have realistic images of objects similar to ImageNet, incorporating co-learning with SHOT and SHOT++ does not affect performance. NRC and AaD performance increase by $0.8\%$ and $0.6\%$, respectively. Several Office-Home and VisDA-C domains contain images of a non-realistic style (e.g. Art, Clipart, Synthetic), and benefit more from co-learning with ImageNet networks. Existing methods improve by $0.5-0.9\%$ on Office-Home, and $0.5-1.7\%$ on VisDA-C. For the challenging DomainNet, SHOT and SHOT++ accuracy are little changed, and AaD accuracy increases by 1.9\%.

\noindent\textbf{Using more robust pre-trained network:}
Co-learning with the stronger and more robust ImageNet Swin-B significantly improves performance in all set-ups tested. Existing methods improve by $2.0-3.3\%$ on Office, $3.6-9.4\%$ on Office-Home, $3.6-4.4\%$ on DomainNet, and $1.4-2.8\%$ on VisDA-C.
Interestingly, on Office and Office-Home, co-learning with ImageNet Swin-B alone is superior to integrating it with the existing SFDA methods tested, which have self-supervised learning components besides pseudolabeling. This implies that effective target domain information in source models is limited, and over-reliance on these models can impede  target adaptation. In contrast, powerful and robust pre-trained networks can have features that are more class-discriminative on target domains, and the large performance boost reflects the advantage of using them to adapt the source models.
\section{Further Analysis}
\label{sec: further analysis}

We conduct further experiments with our proposed strategy in Section~\ref{subsec: further_analysis_colearning} and \ref{subsec: other_adaptation_scenarios}, and discuss considerations on the use of pre-trained networks in Section~\ref{subsec: further_analysis_discussion}.

\subsection{Co-learning with pre-trained networks}
\label{subsec: further_analysis_colearning}

We fix the pre-trained model branch to use ImageNet-1k ConvNeXt-S, which has an intermediate parameter size in our networks tested. Additional experiments on the pre-trained model branch and detailed results are in Appendix.

\noindent\textbf{Adaptation model branch.}
We experiment with other pseudolabeling strategies besides MatchOrConf with a subset of domain pairs from Office-Home dataset in Table~\ref{tab: ablation_pseudolabel_strategy}: SelfConf selects confident samples from adaptation model branch, OtherConf selects confident samples from pre-trained model branch, Match selects samples with same predictions on both branches, and MatchAndConf selects confident samples with same predictions on both branches. SelfConf has the worst performance as source model confidence is not well-calibrated on target domain. Overall, MatchOrConf is the best strategy. From Table~\ref{tab: ablation_confidence}, the optimal confidence threshold $\gamma$ differs across datasets. We estimate the ratio of source to ImageNet feature extractor's target-compatibililty using the ratio of oracle target accuracy, computed by fitting a nearest-centroid-classifier head using fully-labeled target samples for each of the two feature extractors. When the ratio is low as in VisDA-C, the ImageNet feature extractor is more target-compatible, and lowering $\gamma$ allows it to pseudolabel more samples according to its predictions.
In practice, we set $\gamma=0.1$ for VisDA-C as the Real target domain is more similar to ImageNet than to Synthetic source even by visual inspection, and default to $\gamma=0.5$ otherwise although it may not be the optimal value.
\begin{table}

\begin{subtable}[t]{0.48\textwidth}
\centering
 \setlength{\tabcolsep}{2pt}
\begin{adjustbox}{max width=0.75\columnwidth}
\begin{tabular}{l*{4}{c}}
\toprule[1pt]\midrule[0.3pt]
\textbf{Pseudolabel strategy}   & $\mathbf{A\rightarrow C}$ & $\mathbf{A\rightarrow P}$ & $\mathbf{A\rightarrow R}$ & \textbf{Avg} \\ \midrule
SelfConf                        & 49.1  & 74.8  & 77.1  & 67.0 \\
OtherConf                       & 57.5  & \emph{85.4}  & \emph{86.7}  & 76.5 \\
Match                           & \textbf{61.3}  & 84.2  & 86.0  & \emph{77.2} \\
MatchOrConf                     & 59.7  & \textbf{86.3}  & \textbf{87.1}  & \textbf{77.7} \\
MatchAndConf                    & \emph{60.3}  & 81.3  & 84.5  & 75.4 \\
\midrule[0.3pt]\bottomrule[1pt]
\end{tabular}
\end{adjustbox}
\caption{Pseudolabeling strategies, on Office-Home \label{tab: ablation_pseudolabel_strategy}}
\end{subtable}

\begin{subtable}[t]{0.48\textwidth}
\centering
 \setlength{\tabcolsep}{2pt}
\begin{adjustbox}{max width=0.85\columnwidth}
\begin{tabular}{lP{1.5cm}P{2cm}P{1cm}}
\toprule[1pt]\midrule[0.3pt]
\textbf{Confidence threshold $\gamma$}    & \textbf{Office-31}   & \textbf{Office-Home}  & \textbf{VisDA} \\ \midrule
0.1 & 90.5              & 76.1              & \textbf{87.1} \\
0.3 & \textbf{91.2}     & 77.1              & \emph{86.8} \\
0.5 & \emph{91.0}  & \emph{77.4}  & 86.4 \\
0.7 & 90.8              & \textbf{77.5}     & 85.8 \\
0.9 & 90.8              & 77.3              & 85.5 \\ \midrule
Target-compatibility ratio  & 0.982 & 0.947 & 0.844 \\
\midrule[0.3pt]\bottomrule[1pt]
\end{tabular}
\end{adjustbox}
\caption{MatchOrConf confidence threshold. Target-compatibility ratio estimated by source to pre-trained model oracle target accuracy. \label{tab: ablation_confidence}}
\end{subtable}

\vspace{-2mm}
\caption{Co-learning experiments with ImageNet-1k ConvNeXt-S in pre-trained model branch. \label{tab: ablation_adaptation_model_branch}}
\vspace{-4mm}
\end{table}

\begin{figure}
  \centering
  \begin{subfigure}{0.45\linewidth}
    \includegraphics[width=\linewidth]{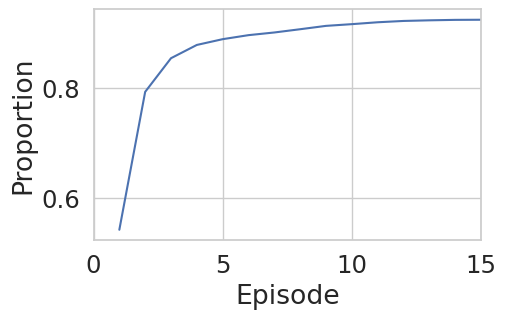}
    \caption{Pseudolabel proportion}
    \label{fig: pseudolabel_proportion}
  \end{subfigure}
  \hfill
  \begin{subfigure}{0.45\linewidth}
    \includegraphics[width=\linewidth]{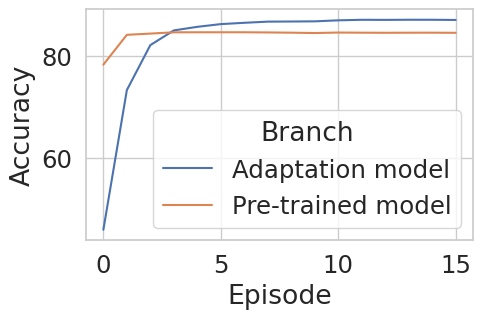}
    \caption{Classification accuracy}
    \label{fig: two_branch_accuracy}
  \end{subfigure}
  
  \vspace{-2mm}
  \caption{VisDA-C co-learning training curves, with ImageNet-1k ConvNeXt-S in pre-trained model branch..}
  \label{fig: training_curves}
  \vspace{-4mm}
\end{figure}

\noindent\textbf{Training curves.} Figure~\ref{fig: training_curves} visualizes the co-learning process for VisDA-C. The proportion of target pseudolabels increases from 0.542 to 0.925 over 15 episodes. Classification accuracy on pre-trained model branch starts higher as the ImageNet feature extractor is more target-compatible than source feature extractor, and flattens earlier since its features are fixed. The adaptation model learns from pre-trained model predictions, and surpasses its accuracy as the feature extractor continues to adapt.

\subsection{Other adaptation scenarios}
\label{subsec: other_adaptation_scenarios}

\noindent \textbf{Non-closed-set settings.}
Our co-learning strategy works well even under label shift where source and target label spaces do not match, as show in Table~\ref{tab: openpartial_officehome_results} on Office-Home. For open-set setting, the first 25 classes are target private, and the next 40 are shared. For partial-set setting, the first 25 classes are shared, and the next 40 are source private. For open-partial setting, the first 10 classes are shared, the next 5 are source private, and the remaining 50 are target private, and we report the harmonic mean of known and unknown class accuracy (H-score) following \cite{yang2022onering}. We add cross-entropy loss with co-learned pseudolabels at temperature 0.1 on open-set and partial-set version of SHOT and open-partial method OneRing \cite{yang2022onering}.

\noindent \textbf{Multi-source adaptation.}
Our proposed strategy can work with multi-source SFDA method as well. In Table~\ref{tab: multisource_results} on Office-31, incorporating co-learned pseudolabels improves performance of CAiDA \cite{dong2021caida}.

\begin{table*}[tb]
\centering
 \setlength{\tabcolsep}{2pt}
\begin{adjustbox}{max width=\textwidth}
\begin{tabular}{l*{12}{c}l}
\toprule[1pt]\midrule[0.3pt]
\textbf{Method}                 & $\mathbf{A\rightarrow C}$ & $\mathbf{A\rightarrow P}$ & $\mathbf{A\rightarrow R}$  
                                & $\mathbf{C\rightarrow A}$ & $\mathbf{C\rightarrow P}$ & $\mathbf{C\rightarrow R}$ 
                                & $\mathbf{P\rightarrow A}$ & $\mathbf{P\rightarrow C}$ & $\mathbf{P\rightarrow R}$
                                & $\mathbf{R\rightarrow A}$ & $\mathbf{R\rightarrow C}$ & $\mathbf{R\rightarrow P}$ & \textbf{Avg}\\ \midrule
Open-set SHOT \cite{liang2020shot}          & 52.8 & 74.5 & 77.1 & \textbf{57.6} & 71.8 & 74.2 & 51.2 & 45.0 & 76.8 & 61.8 & 57.2 & 81.7 & 65.1 \\
\quad w/ Co-learn (w/ ResNet-50)            & \textbf{55.0} & 76.1 & 78.2 & 55.0 & \textbf{73.7} & 73.7 & 53.2 & 47.6 & 77.7 & 61.8 & 57.5 & 82.2 & 66.0 $\green{(\uparrow 0.9)}$ \\
\quad w/ Co-learn (w/ Swin-B)               & 52.2 & \textbf{77.4} & \textbf{78.6} & 56.3 & 71.3 & \textbf{75.3} & \textbf{53.9} & \textbf{49.9} & \textbf{77.9} & \textbf{62.5} & \textbf{58.2} & \textbf{82.3} & \textbf{66.3} $\green{(\uparrow 1.2)}$\\ \hdashline

Partial-set SHOT \cite{liang2020shot}       &  65.9 & \textbf{86.1} & 91.4 & 74.6 & 73.6 & 85.1 & 77.3 & 62.9 & 90.3 & 81.8 & 64.9 & 85.6 & 78.3 \\
\quad w/ Co-learn (w/ ResNet-50)            &  63.9 & 84.7 & 91.8 & 77.6 & 73.6 & 84.5 & 77.4 & 64.8 & \textbf{90.5} & 81.4 & 65.1 & \textbf{87.8} & 78.6 $\green{(\uparrow 0.3)}$ \\
\quad w/ Co-learn (w/ Swin-B)               &  \textbf{66.2} & 85.9 & \textbf{93.2} & \textbf{78.2} & \textbf{74.7} & \textbf{85.6} & \textbf{81.5} & \textbf{65.8} & 90.4 & \textbf{82.6} & \textbf{65.4} & 87.7 & \textbf{79.8} $\green{(\uparrow 1.5)}$ \\ \hdashline

Open-partial OneRing \cite{yang2022onering} & \textbf{68.1} & 81.9 & 87.6 & 72.2 & 76.9 & 83.3 & 80.7 & \textbf{68.8} & 88.2 & 80.5 & 66.0 & 85.5 & 78.3 \\ 
\quad w/ Co-learn (w/ ResNet-50)            & 64.9 & \textbf{86.8} & \textbf{88.0} & 71.9 & 76.7 & 83.5 & \textbf{82.2} & 67.6 & \textbf{89.2} & 82.8 & \textbf{70.0} & 86.4 & \textbf{79.2} $\green{(\uparrow 0.9)}$ \\
\quad w/ Co-learn (w/ Swin-B)               & 65.8 & 85.9 & \textbf{88.0} & \textbf{72.8} & \textbf{77.9} & \textbf{83.9} & 81.9 & 65.8 & 88.5 & \textbf{82.9} & 67.2 & \textbf{87.1} & 79.0 $\green{(\uparrow 0.7)}$ \\
\midrule[0.3pt]\bottomrule[1pt]
\end{tabular}
\end{adjustbox}
\vspace{-2mm}
\caption{Office-Home: Accuracy for open-set and partial-set setting, H-score for open-partial setting, of adapted ResNet-50. \label{tab: openpartial_officehome_results}}
\vspace{-4mm}
\end{table*}
\begin{table}[tb]
\centering
 \setlength{\tabcolsep}{2pt}
\begin{adjustbox}{max width=0.92\columnwidth}
\begin{tabular}{l*{3}{c}l}
\toprule[1pt]\midrule[0.3pt]
\textbf{Method}                 & $\mathbf{\rightarrow A}$ & $\mathbf{\rightarrow D}$ & $\mathbf{\rightarrow W}$ 
                                & \textbf{Avg}\\ \midrule                                
CAiDA \cite{dong2021caida}                  & 75.7 & 98.8 & 93.2 & 89.2  \\ 
\quad w/ Co-learn (w/ ResNet-50)            & 76.8 & 99.0 & 93.2 & 89.7 $\green{(\uparrow 0.5)}$ \\
\quad w/ Co-learn (w/ Swin-B)               & \textbf{79.3} & \textbf{99.6} & \textbf{97.4} & \textbf{92.1} $\green{(\uparrow 2.9)}$\\
\midrule[0.3pt]\bottomrule[1pt]
\end{tabular}
\end{adjustbox}
\vspace{-2mm}
\caption{Multi-source Office-31 accuracy of ResNet-50. \label{tab: multisource_results}}
\vspace{-2mm}
\end{table}

\subsection{Discussion}
\label{subsec: further_analysis_discussion}

We further explore considerations on the characteristics of the pre-trained networks used for co-learning and the effectiveness of our proposed strategy.

\begin{figure}
  \centering
  \begin{subfigure}{0.49\linewidth}
    \includegraphics[width=\linewidth]{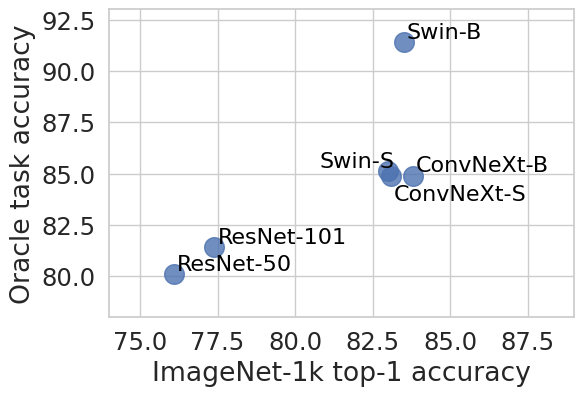}
    \caption{Oracle target accuracy versus ImageNet-1k top-1 accuracy}
  \end{subfigure}
  \hfill
  \begin{subfigure}{0.49\linewidth}
    \includegraphics[width=\linewidth]{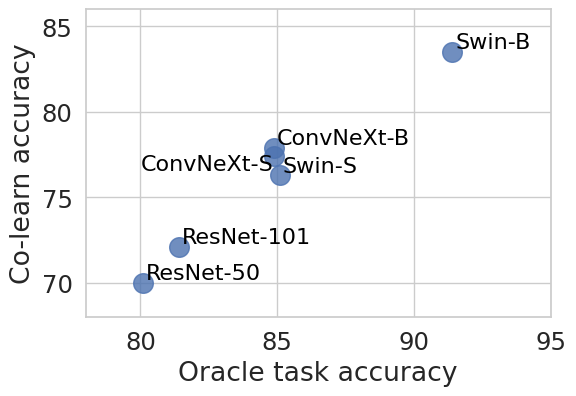}
    \caption{Accuracy after co-learning versus oracle target accuracy}
  \end{subfigure}
  \vspace{-2mm}
  \caption{Evaluations of ImageNet-1k networks. Oracle target accuracy of ImageNet-1k feature extractor, and accuracy of adapted model after co-learning with ImageNet-1k feature extractor, are evaluated on Office-Home.}
  \label{fig: target_compatibility}
  \vspace{-4mm}
\end{figure}

\emph{What are preferred characteristics of the feature extractor for co-learning?} We first study this in Figure~\ref{fig: target_compatibility} through the relationship of ImageNet-1k top-1 accuracy and Office-Home average target domain accuracy. In general, a feature extractor with higher ImageNet accuracy has learned better representations and more robust input-to-feature mappings, and hence is likely to have higher oracle accuracy on target domain and consequently be more helpful in adapting the source model through co-learning. Next, we study a few specific source-domain pairs in Table~\ref{tab: comparison_resnet} by plugging source and ImageNet-1k feature extractors into the pre-trained model branch, and observe effects of the following positive characteristics: (1) Dataset similarity (input style and task): In $C\rightarrow R$ with the same ResNet-50 architecture, there are ImageNet samples more similar to Real World target than Clipart source is, and ImageNet-1k ResNet-50 has higher oracle accuracy than Source ResNet-50. (2) Robustness against covariate shift: In $A\rightarrow C$, although ImageNet is less similar to Clipart target than Art source is, changing from ResNet-50 to ResNet-101 improves oracle accuracy. (3) Different views: In $R\rightarrow A$, although ImageNet-1k ResNet-50 has slightly worse oracle accuracy than Source ResNet-50 ($81.2\%$ vs. $81.8\%$), co-learning with it has better adaptation performance ($71.1\%$ vs. $70.5\%$). Since the adaptation model branch is already initialized by source model, using ImageNet feature extractor in the other branch provides a different view of features and consequently classification decision which benefits co-learning. We refer readers to the Appendix for the full comparison on all 3 datasets.

\begin{table}

\centering
\setlength{\tabcolsep}{2pt}
\begin{adjustbox}{max width=0.97\columnwidth}
\begin{tabular}{l*{4}{c}}
\toprule[1pt]\midrule[0.3pt]
\textbf{Model}  & $\mathbf{A\rightarrow C}$ & $\mathbf{A\rightarrow P}$ & $\mathbf{A\rightarrow R}$ & \textbf{Avg} \\ \midrule
ConvNext-B + classification head$^\dagger$  & 56.1  & 78.6  & 83.3  & 72.7 \\
ResNet-50 adapted w/ ConvNeXt-B             & \textbf{60.5}  & \textbf{86.2}  & \textbf{87.3}  & \textbf{78.0} \\ \midrule
Swin-B + classification head$^\dagger$      & 67.1  & 86.7  & 89.1  & 81.0 \\
ResNet-50 adapted w/ Swin-B                 & \textbf{69.6}  & \textbf{89.6}  & \textbf{91.2}  & \textbf{83.5}\\
\midrule[0.3pt]\bottomrule[1pt]
\end{tabular}
\end{adjustbox}

\caption{Comparison of classification accuracy of ImageNet-1k feature extractor fitted with source-trained classification head, versus classification accuracy of adapted ResNet-50, on Office-Home. $\dagger$ denotes classifier is trained on fully-labeled source data. \label{tab: comparison_two_branch}}
\vspace{-4mm}
\end{table}

\emph{Since modern ImageNet feature extractors have learned better representations, is it sufficient to use them in source models and not adapt?} In Table~\ref{tab: comparison_two_branch}, we fit a 2-layer linear classification head as in Section~ \ref{subsec: experimental setups} on ImageNet-1k ConvNext-B and Swin-B feature extractors by accessing and training the classifier on fully-labeled source data. On the Office-Home domain pairs tested, average performance of Swin-B + classification head is $8.3\%$ higher than ConvNext-B + classification head, showing that feature extractor choice can indeed reduce negative effects of domain shift. However, adaptation to target domain is still necessary. Without having to access the source data and using a larger source model, Source ResNet-50 adapted with our proposed strategy achieves higher classification accuracy.
\section{Conclusion}
\label{sec: conclusion}

In this work, we explored the use of pre-trained networks beyond its current role as initializations in source training in SFDA. We observed that source-training can cause ImageNet networks to lose pre-existing generalization capability on target domain. We proposed distilling useful target domain information from pre-trained networks during target adaptation, and designed a simple co-learning strategy to improve target pseudolabel quality to finetune the source model. This framework also allows us to leverage modern pre-trained networks with improved robustness and generalizability. Experiments on benchmark datasets demonstrate the effectiveness of our framework and strategy.

\section*{Acknowledgments}
This research is supported by the Agency for Science, Technology and Research (A*STAR) under its AME Programmatic Funds (Grant No. A20H6b0151).

{\small
\bibliographystyle{ieee_fullname}
\bibliography{references}
}

\clearpage
\appendix
\section*{Appendix}
\section{Detailed Results}
\label{sec: detailed results}

We provide detailed results of experiments to analyze our proposed framework and strategy.

\subsection{Main results}

In Table~\ref{tab: office31_results_full}, \ref{tab: officehome_results_full} and \ref{tab: visda_results_full}, we provide detailed results of our proposed strategy for the datasets Office-31, Office-Home and VisDA-C with co-learning networks: ResNet-50, ResNet-101, ConvNeXt-S, Swin-S, ConvNeXt-B and Swin-B, where S and B denote the small and base versions of the architectures, respectively. In general, co-learning with the more recently-released ConvNeXt and Swin networks achieves better adaptation performance than co-learning with the ResNets. In particular, co-learning with Swin-B has the best target accuracy in most cases.

In addition, we find that even networks with poor feature extraction ability, such as AlexNet, can provide useful features to improve performance when target style is similar to ImageNet. Co-learned with AlexNet, SHOT and SHOT++ overall Office-Home accuracy (71.9\% and 72.7\%) is little changed at 71.8\%  $\red{(\downarrow 0.1\%)}$ and 72.7\%  $\green{(=)}$. However, 5 out of 12 domain pairs improved by 0.1-1\% and 0.1-1.2\% especially when target is Product or Real World.

\subsection{Further analysis on co-learning with pre-trained networks}

\noindent\textbf{Pre-trained model branch.}
We experiment with different updates to the pre-trained model branch using a subset of domain pairs from Office-Home dataset in Table~\ref{tab: ablation_pretrained_model_branch}. We choose to update no component or different component(s) of the pre-trained network, such as feature extractor, weighted nearest-centroid-classifier (NCC) and a linear 1-layer logit projection layer inserted after NCC. Finetuning just the classifier achieves the best result overall. Co-learning depends on the two branches providing different views of classification decisions. Finetuning the feature extractor or projection layer risks pre-trained model predictions converging too quickly to adaptation model predictions.

\begin{table}[htb]

\centering
 \setlength{\tabcolsep}{2pt}
\begin{adjustbox}{max width=0.95\columnwidth}
\begin{tabular}{P{1.6cm}P{1.5cm}P{1.5cm}*{4}{c}}
\toprule[1pt]\midrule[0.3pt]
\textbf{Feat. extr.} & \textbf{Classifier}    & \textbf{Projection} & $\mathbf{A\rightarrow C}$ & $\mathbf{A\rightarrow P}$ & $\mathbf{A\rightarrow R}$ & \textbf{Avg} \\ \midrule
\xmark      & \xmark        & \xmark                & 59.4  & 83.8  & 86.5  & 76.6 \\
\checkmark  & \xmark        & \xmark                & 59.6  & 82.6  & 86.4  & 76.2 \\
\checkmark  & \checkmark    & \xmark                & \textbf{60.3}  & 84.9  & 86.4  & 77.2 \\
\xmark      & \checkmark    & \xmark                & \emph{59.7}  & \textbf{86.3}  & \emph{87.1}  & \textbf{77.7} \\
\xmark      & \checkmark    & \checkmark            & 59.5  & \emph{85.9}  & \textbf{87.2}  & \emph{77.5} \\
\xmark      & \xmark        & \checkmark            & 59.5  & 84.1  & 86.5  & 76.7 \\
\midrule[0.3pt]\bottomrule[1pt]
\end{tabular}
\end{adjustbox}

\caption{Co-learning experiments on the component finetuned in pre-trained model branch on Office-Home, with ImageNet-1k ConvNeXt-S in pre-trained model branch. \label{tab: ablation_pretrained_model_branch}}
\end{table}
\begin{table}[htb]

\begin{subtable}[t]{0.48\textwidth}
\centering
 \setlength{\tabcolsep}{2pt}
\begin{adjustbox}{max width=0.8\columnwidth}
\begin{tabular}{l*{4}{c}}
\toprule[1pt]\midrule[0.3pt]
\textbf{Centroid computation}   & $\mathbf{A\rightarrow C}$ & $\mathbf{A\rightarrow P}$ & $\mathbf{A\rightarrow R}$ & \textbf{Avg} \\ \midrule
All samples                 & \textbf{59.7}  & \textbf{86.3}  & \textbf{87.1}  & \textbf{77.7} \\
Partial samples             & 59.1  & 86.1  & 87.0  & 77.4 \\
\midrule[0.3pt]\bottomrule[1pt]
\end{tabular}
\end{adjustbox}
\caption{Samples used to estimate class centroids in classifier: all samples versus only samples pseudolabeled by MatchOrConf \label{tab: ablation_ncc_computation}}
\end{subtable}

\begin{subtable}[t]{0.48\textwidth}
\centering
 \setlength{\tabcolsep}{2pt}
\begin{adjustbox}{max width=0.6\columnwidth}
\begin{tabular}{l*{4}{c}}
\toprule[1pt]\midrule[0.3pt]
\textbf{\# Iterations}  & $\mathbf{A\rightarrow C}$ & $\mathbf{A\rightarrow P}$ & $\mathbf{A\rightarrow R}$ & \textbf{Avg}\\ \midrule
1                                           & \textbf{59.7}  & \textbf{86.3}  & 87.1  & \textbf{77.7} \\
2                                           & 58.2  & 85.5  & \textbf{87.2}  & 77.0 \\
3                                           & 57.7  & 85.9  & 86.6  & 76.7 \\
\midrule[0.3pt]\bottomrule[1pt]
\end{tabular}
\end{adjustbox}
\caption{Number of iterations to estimate class centroids in classifier \label{tab: ablation_iteration_ncc_computation}}
\end{subtable}

\caption{Co-learning experiments on the weighted nearest-centroid classifier in pre-trained model branch on Office-Home, with ImageNet-1k ConvNeXt-S in pre-trained model branch. \label{tab: ablation_ncc}}
\end{table}

We also experiment with the computation of weighted nearest-centroid-classifier (NCC) in the pre-trained model branch in Table~\ref{tab: ablation_ncc}. In Table~\ref{tab: ablation_ncc_computation}, we find that class centroids in the NCC classifier should be estimated using all samples, instead of only pseudolabeled samples, to better represent the entire target data distribution. 
Table~\ref{tab: ablation_iteration_ncc_computation} shows that refining class centroids iteratively (as in k-means) is not beneficial, as the model predictions used for centroid estimation have inaccuracies and increased iterations can reinforce the bias that these inaccurate predictions have on the resulting centroids.

\begin{figure*}[htb]
  \centering
  
  \begin{subfigure}{0.29\linewidth}
    \includegraphics[width=\linewidth]{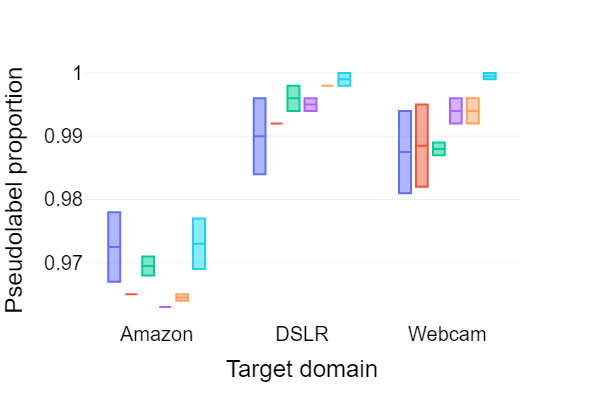}
    \caption{Office-31}
  \end{subfigure}
  \hfill
  \begin{subfigure}{0.39\linewidth}
    \includegraphics[width=\linewidth]{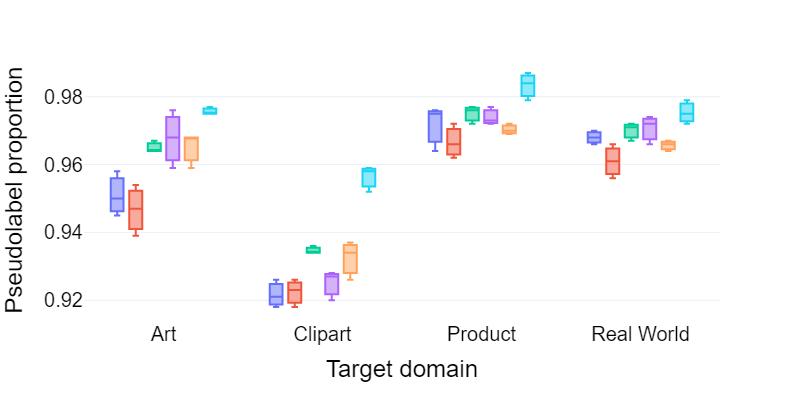}
    \caption{Office-Home}
  \end{subfigure}
  \hfill
  \begin{subfigure}{0.29\linewidth}
    \includegraphics[width=\linewidth]{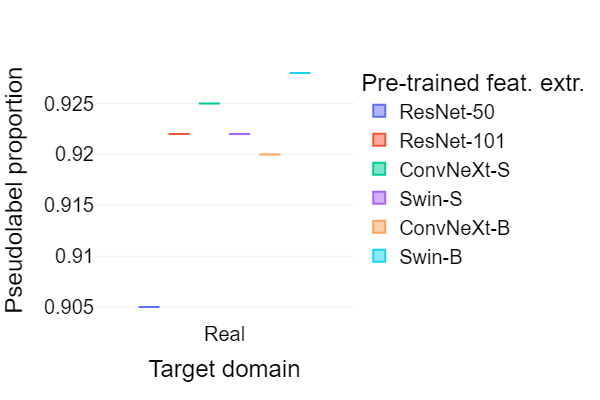}
    \caption{VisDA-C}
  \end{subfigure}  

\caption{Proportion of pseudolabeled samples after co-learning with ImageNet-1k pre-trained feature extractor.}
\label{fig: pseudolabel_proportion_full}
\end{figure*}

\noindent\textbf{Pseudolabel proportion.}
Figure~\ref{fig: pseudolabel_proportion_full} plots the proportion of pseudolabeled samples on target domains after co-learning with different ImageNet pre-trained feature extractors. Overall, the percentage of samples pseudolabeled is $96.4\%$ to $100\%$ in Office-31, $91.8\%$ to $98.7\%$ in Office-Home with percentage above $94\%$ in most cases, and $90.5\%$ to $92.5\%$ in VisDA-C. Samples are not pseudolabeled if the two model branches have the same confidence level and disagree on the class prediction. This disagreement rate is comparatively higher when the target inputs are more different from both source and ImageNet inputs, such as Amazon product images in Office-31 and Clipart images in Office-Home.

\subsection{Further discussion on characteristics of pre-trained networks}

We provide detailed results on the study of preferred feature extractor characteristics for co-learning.

\begin{figure*}[htb]
  \centering
  
  \begin{subfigure}{0.29\linewidth}
    \includegraphics[width=\linewidth]{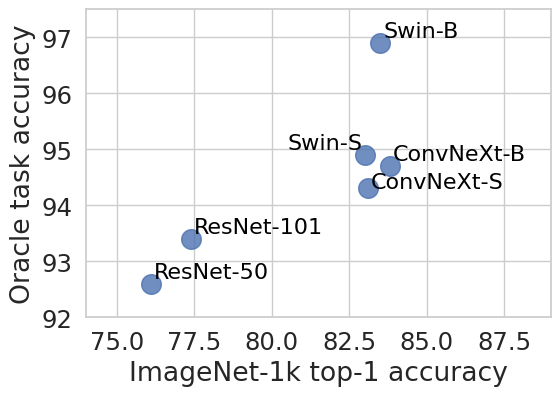}
    \caption{Office-31: Oracle target accuracy versus ImageNet-1k accuracy}
  \end{subfigure}
    \hspace{1mm}
  \begin{subfigure}{0.29\linewidth}
    \includegraphics[width=\linewidth]{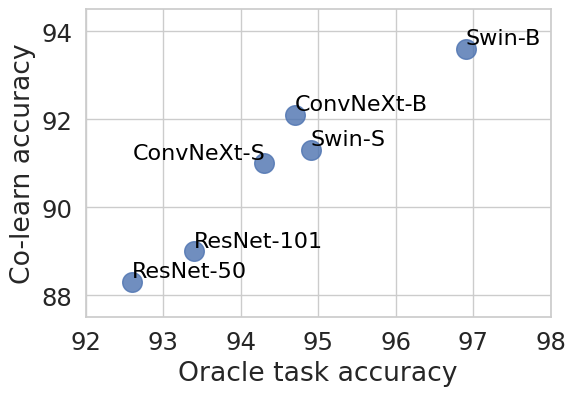}
    \caption{Office-31: Accuracy after co-learning versus oracle target accuracy}
  \end{subfigure}
  
  \begin{subfigure}{0.29\linewidth}
    \includegraphics[width=\linewidth]{figures/officehome_imagenet_vs_oracle.png}
    \caption{Office-Home: Oracle target accuracy versus ImageNet-1k accuracy}
  \end{subfigure}
    \hspace{1mm}
  \begin{subfigure}{0.29\linewidth}
    \includegraphics[width=\linewidth]{figures/officehome_oracle_vs_colearn.png}
    \caption{Office-Home: Accuracy after co-learning versus oracle target accuracy}
  \end{subfigure}  
  
  \begin{subfigure}{0.29\linewidth}
    \includegraphics[width=\linewidth]{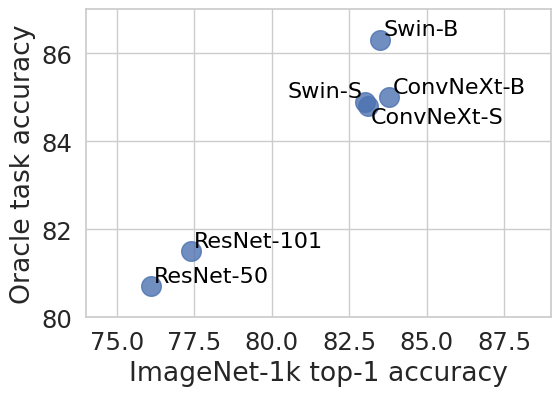}
    \caption{VisDA-C: Oracle target accuracy versus ImageNet-1k accuracy}
  \end{subfigure}
    \hspace{1mm}
  \begin{subfigure}{0.29\linewidth}
    \includegraphics[width=\linewidth]{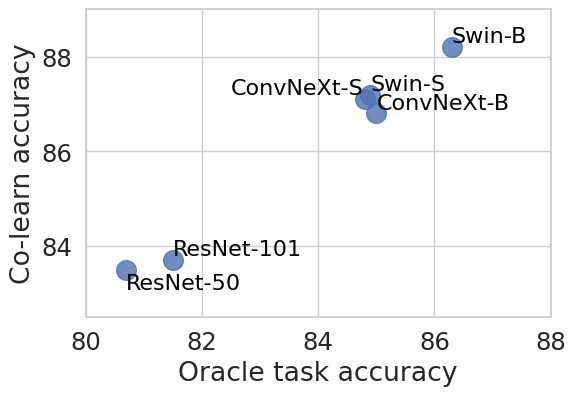}
    \caption{VisDA-C: Accuracy after co-learning versus oracle target accuracy}
  \end{subfigure}  
  
  \caption{Performance evaluations of ImageNet-1k feature extractors. Oracle target accuracy of ImageNet-1k feature extractor, and accuracy of adapted model after co-learning with the feature extractor, are evaluated on dataset target domains.}
  \label{fig: target_compatibility_full}
\end{figure*}

\noindent\textbf{Oracle target accuracy versus ImageNet-1k accuracy.} We plot the relationship of ImageNet-1k top-1 accuracy and average oracle target accuracy attained by pre-trained feature extractors for each dataset in Figure~\ref{fig: target_compatibility_full}, and list the per-domain values in Table~\ref{tab: oracle_full}. Oracle target accuracy is computed by fitting a nearest-centroid-classifier head on each pre-trained feature extractor using fully-labeled target data. In general, higher ImageNet-1k top-1 accuracy corresponds to higher oracle target accuracy. However, there are exceptions. For instance, Swin-B has lower ImageNet-1k accuracy than ConvNext-B ($83.5\%$ vs. $83.8\%$), but higher oracle accuracy for all target domains tested. Despite being trained on the same ImageNet-1k and achieving similar ImageNet-1k performance, the feature extractors learn different features due to differences in architecture and training schemes. While ConvNeXt-B may be more robust to small amounts of input distribution shift due to train-test data splits, Swin-B may be more robust to larger cross-dataset covariate shift and more compatible with the target domains tested.

\noindent\textbf{Adapted model accuracy versus oracle target accuracy} We also plot the relationship of average oracle target accuracy of pre-trained feature extractors and the performance of adapted model after co-learning on target domains in Figure~\ref{fig: target_compatibility_full}. In general, a pre-trained feature extractor with higher oracle target accuracy benefits co-learning as it is more capable of correctly pseudolabeling the target samples, and consequently results in higher adapted model performance. For instance in Table~\ref{tab: comparison_resnet_full}, by using ResNet architectures for co-learning in the pre-trained model branch, we see that the larger and more powerful ImageNet ResNet-101 typically results in higher adapted model performance than ImageNet ResNet-50. There are also exceptions to this correlation. With the same ResNet-50 architecture, although source ResNet-50 has higher oracle accuracy than ImageNet ResNet-50 in some source-target domain pairs (e.g. Office-31 $A\rightarrow D$, $A\rightarrow W$, $W\rightarrow D$; Office-Home $A\rightarrow R$, $P\rightarrow C$, $R\rightarrow A$), the latter results in equal or higher adapted model performance. The ImageNet feature extractors benefit co-learning as they provide an additional view of features and classification decisions different from the source feature extractor already in the adaptation model branch. 

\clearpage

\begin{table*}[htb]

\begin{subtable}[t]{\textwidth}
\centering
\begin{adjustbox}{max width=0.7\textwidth}
\begin{tabular}{l*{7}{c}}
\toprule[1pt]\midrule[0.3pt]
\textbf{Method}                 & \multicolumn{7}{c}{\textbf{Office-31}} \\ \cmidrule(lr){2-8}
                                & $\mathbf{A\rightarrow D}$ & $\mathbf{A\rightarrow W}$ & $\mathbf{D\rightarrow A}$  
                                & $\mathbf{D\rightarrow W}$ & $\mathbf{W\rightarrow A}$ & $\mathbf{W\rightarrow D}$ &\textbf{Avg} \\ \midrule
Source Only                     & 81.9  & 78.0  & 59.4  & 93.6  & 63.4  & 98.8  & 79.2 \\
Co-learn (w/ Resnet-50)         & 93.6  & 90.2  & 75.7  & 98.2  & 72.5  & 99.4  & 88.3\\
Co-learn (w/ Resnet-101)        & 94.2  & 91.6  & 74.7  & 98.6  & 75.6  & 99.6  & 89.0 \\
Co-learn (w/ ConvNeXt-S)        & 96.6  & 92.6  & 79.8  & 97.7  & 79.6  & 99.4  & 91.0\\
Co-learn (w/ Swin-S)            & 96.8  & 93.3  & 79.2  & 98.7  & 80.2  & 99.6  & 91.3\\
Co-learn (w/ ConvNeXt-B)        & \textbf{97.8}  & 96.6  & 80.5  & 98.5  & 79.4  & 99.6  & 92.1\\
Co-learn (w/ Swin-B)            & 97.4  & \textbf{98.2}  & \textbf{84.5}  & \textbf{99.1}  & \textbf{82.2}  & \textbf{100.0} & \textbf{93.6} \\ \hdashline
SHOT$^\dagger$ \cite{liang2020shot}         & 95.0 & 90.4 & 75.2 & \textbf{98.9} & 72.8 & 99.8 & 88.7 \\
\quad w/ Co-learn (w/ ResNet-50)            & 94.2 & 90.2 & 75.7 & 98.2 & 74.4 & \textbf{100.0} & 88.8 \\
\quad w/ Co-learn (w/ ResNet-101)           & \textbf{96.4} & 90.7 & 77.0 & 98.2 & 75.5 & 99.8 & 89.6 \\
\quad w/ Co-learn (w/ ConvNeXt-S)           & 95.2 & 91.4 & 77.5 & \textbf{98.9} & 76.4 & \textbf{100.0} & 89.9 \\
\quad w/ Co-learn (w/ Swin-S)               & 95.2 & 91.6 & 77.7 & 98.5 & \textbf{76.8} & 99.6 & 89.9 \\
\quad w/ Co-learn (w/ ConvNeXt-B)           & 95.8 & 92.8 & 77.2 & 98.4 & 76.4 & \textbf{100.0} & 90.1 \\
\quad w/ Co-learn (w/ Swin-B)               & 95.8 & \textbf{95.6} & \textbf{78.5} & \textbf{98.9} & 76.7 & 99.8 & \textbf{90.9} \\ \hdashline
SHOT++$^\dagger$ \cite{liang2021shotplus}   & 95.6 & 90.8 & 76.0 & 98.2 & 74.6 & \textbf{100.0} & 89.2 \\
\quad w/ Co-learn (w/ ResNet-50)            & 95.0 & 90.7 & 76.4 & 97.7 & 74.9 & 99.8 & 89.1 \\
\quad w/ Co-learn (w/ ResNet-101)           & 95.2 & 90.7 & 77.3 & 98.4 & 76.1 & 99.8 & 89.6\\
\quad w/ Co-learn (w/ ConvNeXt-S)           & 96.0 & 91.8 & 77.1 & 98.4 & 77.1 & \textbf{100.0} & 90.1\\
\quad w/ Co-learn (w/ Swin-S)               & 96.8 & 93.0 & 78.2 & 98.7 & 77.4 & \textbf{100.0} & 90.7 \\
\quad w/ Co-learn (w/ ConvNeXt-B)           & 96.0 & 91.6 & 77.2 & \textbf{99.0} & 76.9 & \textbf{100.0} & 90.1\\
\quad w/ Co-learn (w/ Swin-B)               & \textbf{96.6} & \textbf{93.8} & \textbf{79.8} & 98.9 & \textbf{78.0} & \textbf{100.0} & \textbf{91.2}\\
\midrule[0.3pt]\bottomrule[1pt]
\end{tabular}
\end{adjustbox}
\caption{Office-31: 31-class classification accuracy of adapted ResNet-50 \label{tab: office31_results_full}}
\end{subtable}

\begin{subtable}[t]{\textwidth}
\centering
 \setlength{\tabcolsep}{4pt}
\begin{adjustbox}{max width=\textwidth}
\begin{tabular}{l*{13}{c}}
\toprule[1pt]\midrule[0.3pt]
\textbf{Method}                 & \multicolumn{13}{c}{\textbf{Office-Home}} \\ \cmidrule(lr){2-14}
                                & $\mathbf{A\rightarrow C}$ & $\mathbf{A\rightarrow P}$ & $\mathbf{A\rightarrow R}$  
                                & $\mathbf{C\rightarrow A}$ & $\mathbf{C\rightarrow P}$ & $\mathbf{C\rightarrow R}$ 
                                & $\mathbf{P\rightarrow A}$ & $\mathbf{P\rightarrow C}$ & $\mathbf{P\rightarrow R}$
                                & $\mathbf{R\rightarrow A}$ & $\mathbf{R\rightarrow C}$ & $\mathbf{R\rightarrow P}$ & \textbf{Avg}\\ \midrule
Source Only                     & 43.5  & 67.1  & 74.2  & 51.5  & 62.2  & 63.3  & 51.4  & 40.7  & 73.2  & 64.6  & 45.8  & 77.6  & 59.6 \\
Co-learn (w/ Resnet-50)         & 51.8  & 78.9  & 81.3  & 66.7  & 78.8  & 79.4  & 66.3  & 50.0  & 80.6  & 71.1  & 53.7  & 81.3  & 70.0\\
Co-learn (w/ Resnet-101)        & 54.6  & 81.8  & 83.5  & 68.6  & 79.3  & 80.4  & 68.7  & 52.3  & 82.0  & 72.4  & 57.1  & 84.1  & 72.1 \\
Co-learn (w/ ConvNeXt-S)        & 59.7  & 86.3  & 87.1  & 75.9  & 84.5  & 86.8  & 76.1  & 58.7  & 87.1  & 78.0  & 61.9  & 87.2  & 77.4 \\
Co-learn (w/ Swin-S)            & 56.4  & 85.1  & 88.0  & 73.9  & 83.7  & 86.1  & 75.4  & 55.3  & 87.8  & 77.3  & 58.9  & 87.9  & 76.3 \\
Co-learn (w/ ConvNeXt-B)        & 60.5  & 85.9  & 87.2  & 76.1  & 85.3  & 86.6  & 76.5  & 58.6  & 87.5  & 78.9  & 62.4  & 88.8  & 77.9 \\
Co-learn (w/ Swin-B)            & \textbf{69.6}  & \textbf{89.5}  & \textbf{91.2}  & \textbf{82.7}  & \textbf{88.4}  & \textbf{91.3}  & \textbf{82.6}  & \textbf{68.5}  & \textbf{91.5}  & \textbf{82.8}  & \textbf{71.3}  & \textbf{92.1}  & \textbf{83.5} \\ \hdashline
SHOT$^\dagger$ \cite{liang2020shot}         & 55.8 & 79.6 & 82.0 & 67.4 & 77.9 & 77.9 & 67.6 & 55.6 & 81.9 & 73.3 & 59.5 & 84.0 & 71.9 \\
\quad w/ Co-learn (w/ ResNet-50)            & 56.3 & 79.9 & 82.9 & 68.5 & 79.6 & 78.7 & 68.1 & 54.8 & 82.5 & 74.5 & 59.0 & 83.6 & 72.4 \\
\quad w/ Co-learn (w/ ResNet-101)           & 57.2 & 80.6 & 83.5 & 69.1 & 79.1 & 79.9 & 69.1 & 56.5 & 82.3 & 74.8 & 59.9 & 85.3 & 73.1 \\
\quad w/ Co-learn (w/ ConvNeXt-S)           & 58.9 & 81.5 & 84.6 & 71.7 & 79.7 & 82.0 & 71.3 & 57.5 & 84.2 & 75.9 & 61.9 & 86.0 & 74.6 \\
\quad w/ Co-learn (w/ Swin-S)               & 58.9 & 81.3 & 84.5 & 70.7 & 80.1 & 81.9 & 69.8 & 57.8 & 83.7 & 75.4 & 61.0 & 86.0 & 74.3\\
\quad w/ Co-learn (w/ ConvNeXt-B)           & 60.0 & 81.1 & 84.3 & 71.4 & 80.4 & 81.5 & 70.9 & 58.8 & 83.8 & \textbf{76.2} & 62.8 & 86.3 & 74.8\\
\quad w/ Co-learn (w/ Swin-B)               & \textbf{61.7} & \textbf{82.9} & \textbf{85.3} & \textbf{72.7} & \textbf{80.5} & \textbf{82.0} & \textbf{71.6} & \textbf{60.4} & \textbf{84.5} & 76.0 & \textbf{64.3} & \textbf{86.7} & \textbf{75.7}\\ \hdashline
SHOT++$^\dagger$ \cite{liang2021shotplus}   & 57.1 & 79.5 & 82.6 & 68.5 & 79.5 & 78.6 & 68.3 & 56.1 & 82.9 & 74.0 & 59.8 & 85.0 & 72.7 \\
\quad w/ Co-learn (w/ ResNet-50)            & 57.7 & 81.1 & 84.0 & 69.2 & 79.8 & 79.2 & 69.1 & 57.7 & 82.9 & 73.7 & 60.1 & 85.0 & 73.3 \\
\quad w/ Co-learn (w/ ResNet-101)           & 59.4 & 80.5 & 84.0 & 69.3 & 80.3 & 80.5 & 69.8 & 57.7 & 83.2 & 74.2 & 60.4 & 85.6 & 73.7\\
\quad w/ Co-learn (w/ ConvNeXt-S)           & 60.7 & 82.4 & 84.9 & 71.1 & 80.7 & 82.3 & 70.8 & 59.4 & 84.2 & 75.8 & 63.2 & 85.9 & 75.1\\
\quad w/ Co-learn (w/ Swin-S)               & 60.4 & 81.6 & 84.7 & 71.0 & 80.7 & 81.3 & 71.1 & 57.4 & 84.6 & 75.7 & 61.9 & 85.8 & 74.7\\
\quad w/ Co-learn (w/ ConvNeXt-B)           & 60.8 & 80.9 & 84.5 & 70.8 & 81.2 & 83.1 & 70.9 & \textbf{60.1} & 84.3 & \textbf{76.4} & 62.5 & 85.6 & 75.1\\
\quad w/ Co-learn (w/ Swin-B)               & \textbf{63.7} & \textbf{83.0} & \textbf{85.7} & \textbf{72.6} & \textbf{81.5} & \textbf{83.8} & \textbf{72.0} & 59.9 & \textbf{85.3} & 76.3 & \textbf{65.3} & \textbf{86.6} & \textbf{76.3} \\
\midrule[0.3pt]\bottomrule[1pt]
\end{tabular}
\end{adjustbox}
\caption{Office-Home: 65-class classification accuracy of adapted ResNet-50 \label{tab: officehome_results_full}}
\end{subtable}
\end{table*}

\pagebreak

\begin{table*}[htb]
\ContinuedFloat
\begin{subtable}[t]{\textwidth}
\centering
\begin{adjustbox}{max width=0.9\textwidth}
\begin{tabular}{l*{13}{c}}
\toprule[1pt]\midrule[0.3pt]
\textbf{Method}                 & \multicolumn{13}{c}{\textbf{VisDA-C}} \\ \cmidrule(lr){2-14}
                                & \textbf{plane} & \textbf{bike} & \textbf{bus} & \textbf{car} & \textbf{horse} & \textbf{knife} & \textbf{mcycle} & \textbf{person} & \textbf{plant} & \textbf{sktbrd} & \textbf{train} & \textbf{truck} & \textbf{Avg}\\ \midrule
Source Only                     & 51.5  & 15.3  & 43.4  & 75.4  & 71.2  & 6.8   & 85.5  & 18.8  & 49.4  & 46.4  & 82.1  & 5.4   & 45.9 \\
Co-learn (w/ Resnet-50)         & 96.2  & 76.2  & 77.5  & 77.8  & 93.8  & 96.6  & 91.5  & 76.7  & 90.4  & 90.8  & 86.0  & 48.9  & 83.5 \\
Co-learn (w/ Resnet-101)        & 96.5  & 78.9  & 77.5  & 75.7  & 94.6  & 95.8  & 89.1  & 77.7  & 90.5  & 91.0  & 86.2  & 51.5  & 83.7 \\
Co-learn (w/ ConvNeXt-S)        & 97.8  & 89.7  & 82.3  & \textbf{81.3}  & 97.3  & 97.8  & 93.4  & 66.9  & 95.4  & 96.0  & 90.7  & \textbf{56.5}  & 87.1 \\
Co-learn (w/ Swin-S)            & 97.8  & 88.5  & 84.7  & 78.5  & 96.8  & 97.8  & 93.3  & 73.9  & 94.9  & 94.8  & 91.2  & 54.8  & 87.2 \\
Co-learn (w/ ConvNeXt-B)        & 98.0  & 89.2  & \textbf{84.9}  & 80.2  & 97.0  & 98.4  & 93.6  & 64.3  & \textbf{95.6}  & \textbf{96.3}  & 90.4  & 54.0  & 86.8 \\
Co-learn (w/ Swin-B)            & \textbf{99.0}  & \textbf{90.0}  & 84.2  & 81.0  & \textbf{98.1}  & \textbf{97.9}  & \textbf{94.9}  & \textbf{80.1}  & 94.8  & 95.9  & \textbf{94.4}  & 48.1  & \textbf{88.2} \\ \hdashline
SHOT$^\dagger$ \cite{liang2020shot}         & 95.3 & 87.1 & 79.1 & 55.1 & 93.2 & 95.5 & 79.5 & 79.6 & 91.6 & 89.5 & 87.9 & 56.0 & 82.4 \\
\quad w/ Co-learn (w/ ResNet-50)            & 95.5 & 86.1 & 76.3 & 57.6 & 93.2 & \textbf{96.9} & 83.4 & 81.1 & 92.0 & 90.5 & 84.5 & 60.7 & 83.1 \\
\quad w/ Co-learn (w/ ResNet-101)           & 94.9 & 84.8 & 77.7 & 63.0 & 94.1 & 95.6 & 85.6 & 81.0 & 93.0 & 92.2 & 86.4 & 60.4 & 84.1 \\
\quad w/ Co-learn (w/ ConvNeXt-S)           & 95.6 & \textbf{89.8} & \textbf{82.2} & \textbf{66.2} & 94.2 & 96.3 & 85.4 & 82.0 & 92.8 & 91.5 & 86.6 & 59.3 & 85.1 \\
\quad w/ Co-learn (w/ Swin-S)               & 95.2 & 88.1 & 81.5 & 63.9 & 93.9 & 95.6 & 86.7 & 82.1 & \textbf{94.0} & 92.3 & 87.3 & \textbf{63.9} & \textbf{85.3} \\
\quad w/ Co-learn (w/ ConvNeXt-B)           & 95.1 & 87.5 & 81.2 & 62.4 & \textbf{94.4} & 96.4 & 86.2 & \textbf{82.3} & 93.1 & \textbf{92.5} & 88.5 & 63.0 & 85.2 \\
\quad w/ Co-learn (w/ Swin-B)               & \textbf{96.0} & 88.1 & 81.0 & 63.0 & 94.3 & 95.9 & \textbf{87.1} & 81.8 & 92.8 & 91.9 & \textbf{90.1} & 60.5 & 85.2 \\ \hdashline
SHOT++$^\dagger$ \cite{liang2021shotplus}   & 94.5 & 88.5 & \textbf{90.4} & 84.6 & \textbf{97.9} & \textbf{98.6} & 91.9 & 81.8 & 96.7 & 91.5 & 93.8 & 31.3 & 86.8 \\
\quad w/ Co-learn (w/ ResNet-50)            & 97.5 & 89.8 & 86.4 & 82.7 & 97.8 & 98.5 & 92.7 & 83.5 & 97.3 & 90.8 & 97.5 & 36.7 & 87.4 \\
\quad w/ Co-learn (w/ ResNet-101)           & 97.9 & 88.6 & 86.8 & \textbf{86.7} & \textbf{97.9} & \textbf{98.6} & 92.4 & 83.6 & 97.4 & 92.5 & 94.4 & 32.5 & 87.4 \\
\quad w/ Co-learn (w/ ConvNeXt-S)           & 97.5 & 90.3 & 89.1 & 85.6 & 97.7 & 98.1 & \textbf{93.2} & 84.7 & 96.5 & 92.6 & \textbf{95.1} & 39.1 & 88.3 \\
\quad w/ Co-learn (w/ Swin-S)               & 97.6 & 87.3 & 86.8 & 83.3 & \textbf{97.9} & 98.2 & 92.1 & 84.4 & 97.6 & 90.2 & 94.7 & 42.0 & 87.7 \\
\quad w/ Co-learn (w/ ConvNeXt-B)           & 97.7 & 88.6 & 86.2 & \textbf{86.7} & \textbf{97.9} & 98.3 & 90.8 & 82.8 & \textbf{97.7} & 90.3 & 95.0 & 39.0 & 87.6 \\
\quad w/ Co-learn (w/ Swin-B)               & \textbf{98.0} & \textbf{91.1} & 88.6 & 83.2 & 97.8 & 97.8 & 92.0 & \textbf{85.8} & 97.6 & \textbf{93.2} & 95.0 & \textbf{43.5} & \textbf{88.6} \\
\midrule[0.3pt]\bottomrule[1pt]
\end{tabular}
\end{adjustbox}
\caption{VisDA-C: 12-class classification accuracy of adapted ResNet-101 \label{tab: visda_results_full}}
\end{subtable}

\caption{Classification accuracy of adapted models. For proposed strategy, ImageNet-1k feature extractor used is given in parenthesis. $\dagger$ denotes reproduced results. \label{tab: results_full}}
\end{table*}
\begin{table*}[htb]

\begin{subtable}[t]{\textwidth}
\centering
\begin{adjustbox}{max width=0.75\textwidth}
\begin{tabular}{ll*{6}{c}}
\toprule[1pt]\midrule[0.3pt]
\textbf{Pre-training data} & \textbf{Feat. extr.}  & \textbf{\# Param.} & \textbf{IN-1k Acc@1}  & \multicolumn{4}{c}{\textbf{Office-31}} \\ \cmidrule(lr){5-8}
& & & & \textbf{Amazon}  & \textbf{DSLR}  &  \textbf{Webcam}    &  \textbf{Avg} \\ \midrule
ImageNet-1k, then Amazon   & ResNet-50     & 26M   & -     &  -    & 98.6  & 98.1  & -  \\
ImageNet-1k, then DSLR     & ResNet-50     & 26M   & -     & 80.1  & -     & 97.7  & -  \\
ImageNet-1k, then Webcam   & ResNet-50     & 26M   & -     & 81.7  & 99.2  & -     & -    \\ \hdashline
ImageNet-1k                & ResNet-50     & 26M   & 76.1  & 81.8  & 98.2  & 97.9  & 92.6 \\
ImageNet-1k                & ResNet-101    & 45M   & 77.4  & 83.4  & 98.8  & 97.9  & 93.4 \\
ImageNet-1k                & ConvNeXt-S    & 50M   & 83.1  & 85.2  & \underline{99.4}  & 98.2  & 94.3 \\
ImageNet-1k                & Swin-S        & 50M   & 83.0  & \underline{86.2}  & 99.0  & \underline{99.4}  & \underline{94.9} \\
ImageNet-1k                & ConvNeXt-B    & 89M   & \textbf{83.8}  & \underline{86.2}  & \underline{99.4}  & 98.6  & 94.7 \\
ImageNet-1k                & Swin-B        & 88M   & \underline{83.5}  & \textbf{90.6}  & \textbf{100.0} & \textbf{100.0} & \textbf{96.9} \\ \midrule
\midrule[0.3pt]\bottomrule[1pt]
\end{tabular}
\end{adjustbox}
\caption{Oracle target accuracy of pre-trained feature extractors on Office-31 \label{tab: office31_oracle}}
\end{subtable}

\begin{subtable}[t]{\textwidth}
\centering
\begin{adjustbox}{max width=0.9\textwidth}
\begin{tabular}{llccP{1.2cm}P{1.2cm}P{1.2cm}P{2cm}P{1.2cm}}
\toprule[1pt]\midrule[0.3pt]
\textbf{Pre-training data} & \textbf{Feat. extr.} & \textbf{\# Param.} & \textbf{IN-1k Acc@1} & \multicolumn{5}{c}{\textbf{Office-Home}} \\ \cmidrule(lr){5-9}
& & & & \textbf{Art}  & \textbf{Clipart}  &  \textbf{Product}    & \textbf{Real World}  & \textbf{Avg} \\ \midrule
ImageNet-1k, then Art          & ResNet-50     & 26M   & -     & -     & 69.5  & 88.1  & 86.3  & - \\
ImageNet-1k, then Clipart      & ResNe-50      & 26M   & -     & 79.3  & -     & 86.0  & 83.3  & - \\
ImageNet-1k, then Product      & ResNet-50     & 26M   & -     & 80.0  & 67.5  & -     & 85.8  & - \\
ImageNet-1k, then Real World   & ResNet-50     & 26M   & -     & 81.8  & 69.1  & 88.1  & -     & - \\ \hdashline
ImageNet-1k                    & ResNet-50     & 26M   & 76.1  & 81.2  & 65.5  & 87.7  & 86.0  & 80.1 \\
ImageNet-1k                    & ResNet-101    & 45M   & 77.4  & 82.5  & 68.0  & 88.2  & 87.0  & 81.4 \\
ImageNet-1k                    & ConvNeXt-S    & 50M   & 83.1  & 86.0  & 72.3  & \underline{91.7}  & 89.4  & 84.9 \\
ImageNet-1k                    & Swin-S        & 50M   & 83.0  & \underline{86.6}  & 72.0  & 91.5  & \underline{90.2}  & \underline{85.1} \\
ImageNet-1k                    & ConvNeXt-B    & 89M   & \textbf{83.8}  & 85.7  & \underline{73.4}  & 91.4  & 89.2  & 84.9 \\
ImageNet-1k                    & Swin-B        & 88M   & \underline{83.5}  & \textbf{92.3}  & \textbf{83.1}  & \textbf{95.3}  & \textbf{94.7}  & \textbf{91.4} \\ \midrule
\midrule[0.3pt]\bottomrule[1pt]
\end{tabular}
\end{adjustbox}
\caption{Oracle target accuracy of pre-trained feature extractors on Office-Home \label{tab: officehome_oracle}}
\end{subtable}

\begin{subtable}[t]{\textwidth}
\centering
\begin{adjustbox}{max width=0.6\textwidth}
\begin{tabular}{ll*{3}{c}}
\toprule[1pt]\midrule[0.3pt]
\textbf{Pre-training data} & \textbf{Feat. extr.} & \textbf{\# Param.} & \textbf{IN-1k Acc@1} & \textbf{VisDA-C} \\ \midrule
ImageNet-1k, then Synthetic& ResNet-101    & 45M   & -     & 71.6 \\ \hdashline
ImageNet-1k                & ResNet-50     & 26M   & 76.1  & 80.7 \\
ImageNet-1k                & ResNet-101    & 45M   & 77.4  & 81.5 \\
ImageNet-1k                & ConvNeXt-S    & 50M   & 83.1  & 84.8 \\
ImageNet-1k                & Swin-S        & 50M   & 83.0  & 84.9 \\
ImageNet-1k                & ConvNeXt-B    & 89M   & \textbf{83.8}  & \underline{85.0} \\
ImageNet-1k                & Swin-B        & 88M   & \underline{83.5}  & \textbf{86.3} \\ \midrule
\midrule[0.3pt]\bottomrule[1pt]
\end{tabular}
\end{adjustbox}
\caption{Oracle target accuracy of pre-trained feature extractors on VisDA-C \label{tab: visda_oracle}}
\end{subtable}

\caption{Oracle target accuracy of pre-trained feature extractors fitted with nearest-centroid-classifier heads learned on fully-labeled target domain data. IN-1k Acc@1 denotes ImageNet-1k top-1 accuracy. Feature extractors above dash line are from source models, below the dash line are from ImageNet-1k networks. \label{tab: oracle_full}}
\end{table*}
\begin{table*}[htb]

\begin{subtable}[t]{\textwidth}
\centering
\begin{adjustbox}{max width=0.65\textwidth}
\begin{tabular}{l*{7}{c}}
\toprule[1pt]\midrule[0.3pt]
\textbf{Pre-trained feat. extr.}   & \multicolumn{7}{c}{\textbf{Office-31}} \\ \cmidrule(lr){2-8}
                                & $\mathbf{A\rightarrow D}$ & $\mathbf{A\rightarrow W}$ & $\mathbf{D\rightarrow A}$  
                                & $\mathbf{D\rightarrow W}$ & $\mathbf{W\rightarrow A}$ & $\mathbf{W\rightarrow D}$ &\textbf{Avg} \\ \midrule
Source ResNet-50        & 92.6  & 89.9  & 74.0  & 96.1  & \emph{73.9}  & \emph{99.4}  & 87.6\\
ImageNet-1k ResNet-50   & \emph{93.6}  & \emph{90.2}  & \textbf{75.7}  & \emph{98.2}  & 72.5  & \emph{99.4}  & \emph{88.3} \\
ImageNet-1k ResNet-101  & \textbf{94.2}  & \textbf{91.6}  & \emph{74.7}  & \textbf{98.6}  & \textbf{75.6}  & \textbf{99.6}  & \textbf{89.0} \\
\midrule[0.3pt]\bottomrule[1pt]
\end{tabular}
\end{adjustbox}
\caption{Classification accuracy of adapted ResNet-50 on Office-31 \label{tab: office31_comparison_resnet}}
\end{subtable}

\begin{subtable}[t]{\textwidth}
\centering
\begin{adjustbox}{max width=\textwidth}
\begin{tabular}{l*{13}{c}}
\toprule[1pt]\midrule[0.3pt]
\textbf{Pre-trained feat. extr.}   & \multicolumn{13}{c}{\textbf{Office-Home}} \\ \cmidrule(lr){2-14}
                                & $\mathbf{A\rightarrow C}$ & $\mathbf{A\rightarrow P}$ & $\mathbf{A\rightarrow R}$  
                                & $\mathbf{C\rightarrow A}$ & $\mathbf{C\rightarrow P}$ & $\mathbf{C\rightarrow R}$ 
                                & $\mathbf{P\rightarrow A}$ & $\mathbf{P\rightarrow C}$ & $\mathbf{P\rightarrow R}$
                                & $\mathbf{R\rightarrow A}$ & $\mathbf{R\rightarrow C}$ & $\mathbf{R\rightarrow P}$ & \textbf{Avg}\\ \midrule
Source ResNet-50        & \emph{53.9}  & \emph{79.0}  & \emph{81.3}  & 63.5  & 75.6  & 74.1  & 63.7  & 49.8  & 80.1  & 70.5  & \emph{54.8}  & \emph{81.4}  & 69.0\\
ImageNet-1k ResNet-50   & 51.8  & 78.9  & \emph{81.3}  & \emph{66.7}  & \emph{78.8}  & \emph{79.4}  & \emph{66.3}  & \emph{50.0}  & \emph{80.6}  & \emph{71.1}  & 53.7  & 81.3  & \emph{70.0} \\
ImageNet-1k ResNet-101  & \textbf{54.6}  & \textbf{81.8}  & \textbf{83.5}  & \textbf{68.6}  & \textbf{79.3}  & \textbf{80.4}  & \textbf{68.7}  & \textbf{52.3}  & \textbf{82.0}  & \textbf{72.4}  & \textbf{57.1}  & \textbf{84.1}  & \textbf{72.1} \\
\midrule[0.3pt]\bottomrule[1pt]
\end{tabular}
\end{adjustbox}
\caption{Classification accuracy of adapted ResNet-50 on Office-Home \label{tab: officehome_comparison_resnet}}
\end{subtable}

\begin{subtable}[t]{\textwidth}
\centering
\begin{adjustbox}{max width=0.9\textwidth}
\begin{tabular}{l*{13}{c}}
\toprule[1pt]\midrule[0.3pt]
\textbf{Pre-trained feat. extr.}   & \multicolumn{13}{c}{\textbf{VisDA-C}} \\ \cmidrule(lr){2-14}
                                & \textbf{plane} & \textbf{bike} & \textbf{bus} & \textbf{car} & \textbf{horse} & \textbf{knife} & \textbf{mcycle} & \textbf{person} & \textbf{plant} & \textbf{sktbrd} & \textbf{train} & \textbf{truck} & \textbf{Avg}\\ \midrule
Source ResNet-101       & 63.6  & 55.6  & \emph{68.0}  & 65.6  & 84.0  & 95.6  & 89.0  & 65.8  & 82.6  & 4.9     & 82.2  & 25.8  & 65.2 \\
ImageNet-1k ResNet-50   & \emph{96.2}  & \emph{76.2}  & \textbf{77.5}  & \textbf{77.8}  & \emph{93.8}  & \textbf{96.6}  & \textbf{91.5}  & \emph{76.7}  & \emph{90.4}  & \emph{90.8}    & \emph{86.0}  & \emph{48.9}  & \emph{83.5} \\
ImageNet-1k ResNet-101  & \textbf{96.5}  & \textbf{78.9}  & \textbf{77.5}  & \emph{75.7}  & \textbf{94.6}  & \emph{95.8}  & \emph{89.1}  & \textbf{77.7}  & \textbf{90.5}  & \textbf{91.0}    & \textbf{86.2}  & \textbf{51.5}  & \textbf{83.7} \\
\midrule[0.3pt]\bottomrule[1pt]
\end{tabular}
\end{adjustbox}
\caption{Classification accuracy of adapted ResNet-101 on VisDA-C \label{tab: visda_comparison_resnet}}
\end{subtable}

\caption{Classification accuracy of adapted model after co-learning with ResNets. \label{tab: comparison_resnet_full}}
\end{table*}

\end{document}